\documentclass[10pt,twocolumn,letterpaper]{article}

\usepackage{cvpr}              % To produce the CAMERA-READY version
\usepackage{booktabs}
\usepackage{siunitx}
\usepackage{xcolor}  %也可以使用{color}宏包，但是颜色库里面颜色较少，使用xcolor会更好
\usepackage{tcolorbox}
\usepackage{graphicx}
\usepackage{caption}

\usepackage{lipsum}
\usepackage{cvpr} % or iccv
\usepackage{siunitx}
\usepackage{xcolor}
\usepackage{colortbl, booktabs}

\usepackage{float}
\usepackage{algorithm,algpseudocode} % 提供浮动体环境，用于包含算法并添加标题和标签
\usepackage{amsmath} % 用于排版数学公式，伪代码中经常用到
\definecolor{cvprblue}{rgb}{0.21,0.49,0.74}
\usepackage[pagebackref,breaklinks,colorlinks,citecolor=cvprblue]{hyperref}
\def\paperID{*****} % *** Enter the Paper ID here
\def\confName{CVPR}
\def\confYear{2026}

\title{SafeRoPE: Risk-specific Head-wise Embedding Rotation for Safe Generation in Rectified Flow Transformers}

\author{
Xiang Yang$^{1}$ \quad
Feifei Li$^{1}$ \quad
Mi Zhang$^{1\dag}$\quad
Geng Hong$^{1\dag}$\quad
Xiaoyu You$^{2}$ \quad
Min Yang$^{1}$ \\
$^{1}$Fudan University, Shanghai, China \\
$^{2}$East China University of Science and Technology, Shanghai, China \\
{\tt\small
$^{1}$\{yangx25@m., ffli23@m., mi\_zhang@, ghong@, m\_yang@\}fudan.edu.cn, \quad
$^{2}${\tt\small xiaoyuyou@ecust.edu.cn}
}
}
\newcommand{\ours}{SafeRoPE }
\newcommand{\dev}{FLUX.1-dev}
\newcommand{\sch}{FLUX.1-sch}
\begin{document}
% \label{sec:intro}

\twocolumn[{
\renewcommand\twocolumn[1][]{#1}
\maketitle
\begin{center}
    \captionsetup{type=figure}
    \includegraphics[width=\textwidth]{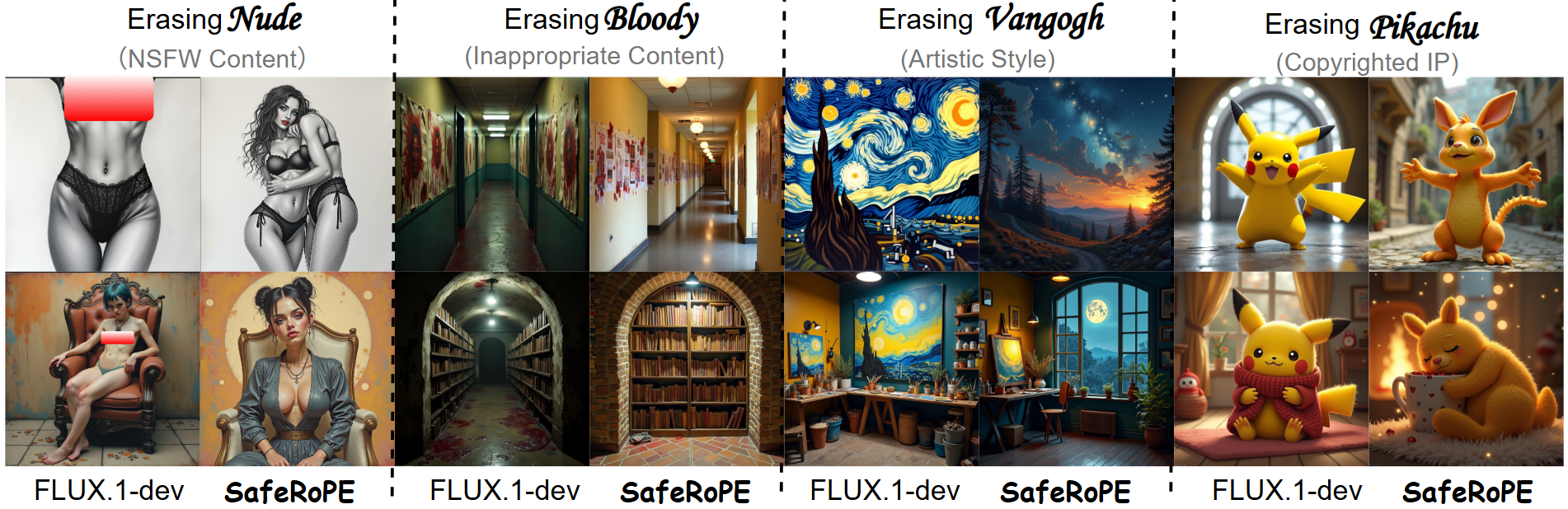}
    \captionof{figure}{Qualitative comparison of concept erasure on FLUX.1-dev. By performing risk-specific rotations, SafeRoPE effectively suppresses various undesired concepts, while maintaining high visual quality and semantic fidelity.}
    \label{teaser}
\end{center}
}]
%     \centering
%     \includegraphics[width=0.9\textwidth]{sec/Figs/intro1_wx.png}
%     \caption{Illustration of SafeRoPE framework: a head-wise, risk-aware method for safety alignment in rectified flow transformers.}
% \maketitlesupplementary    % 只出标题

% 现在放 teaser，使用浮动体（跨双栏）
\footnotetext[2]{Corresponding author}

\begin{abstract}
Recent Text-to-Image (T2I) models based on rectified-flow transformers (e.g., SD3, FLUX) achieve high generative fidelity but remain vulnerable to unsafe semantics, especially when triggered by multi-token interactions. Existing mitigation methods largely rely on fine-tuning or attention modulation for concept unlearning; however, their expensive computational overhead and design tailored to U-Net-based denoisers hinder direct adaptation to transformer-based diffusion models (e.g., MMDiT). 
In this paper, we conduct an in-depth analysis of the attention mechanism in MMDiT and find that unsafe semantics concentrate within interpretable, low-dimensional subspaces at head level, where a finite set of \textbf{safety-critical heads} is responsible for unsafe feature extraction. We further observe that perturbing the Rotary Positional Embedding (RoPE) applied to the query and key vectors can effectively modify some specific concepts in the generated images.
Motivated by these insights, we propose SafeRoPE, a lightweight and fine-grained safe generation framework for MMDiT. 
Specifically, SafeRoPE first constructs head-wise unsafe subspaces by decomposing unsafe embeddings within safety-critical heads, and computes a Latent Risk Score (LRS) for each input vector via projection onto these subspaces. We then introduce head-wise RoPE perturbations that can suppress unsafe semantics without degrading benign content or image quality.
% , which can be interpreted as an additional rotation for token embedding. % that can suppress unsafe semantics %while preserving benign ones without corrupt the image quality.
% LRS: norm of embedding that projected to unsafe subspace  
% Since RoPE operates as a rotation in the embedding space, 
SafeRoPE combines both head-wise LRS and RoPE perturbations to perform risk-specific head-wise rotation on query and key vector embeddings, enabling precise suppression of unsafe outputs while maintaining generation fidelity. Extensive experiments demonstrate that SafeRoPE achieves SOTA performance in balancing effective harmful content mitigation and utility preservation for safe generation of MMDiT. \textit{\textbf{Codes are available at https://github.com/deng12yx/SafeRoPE.}}
 % Moreover, the overall process requires no fine-tuning and therefore introduces no loss to the model’s generation diversity
% and parameter redundancy exists within individual heads; ($ii$) randomly perturbing text position IDs in RoPE (the positional encoding of FLUX) can suppress unsafe outputs while preserving normal semantics. Motivated by these findings, we propose SafeRoPE, a training-free, risk-aware embedding rotation framework that leverages RoPE’s geometry in rectified-flow transformers for fine-grained safety modulation. SafeRoPE first performs singular value decomposition (SVD) on the unsafe feature space within each head to evaluate its unsafe extraction capacity, and computes a continuous Latent Risk Score(LRS) for each input vector via projection. Second, it integrates LRS with RoPE-based rotations, applying controlled head-wise rotations within the unsafe subspace, thereby enables precise suppression of unsafe semantics while preserving overall fidelity. Extensive experiments demonstrate that SafeRoPE effectively mitigates unsafe generation outputs without compromising image quality compared with the SOTA method. Our findings reveal that unsafe semantics concentrate within interpretable, low-dimensional RoPE subspaces, establishing RoPE as a principled mechanism for safety alignment in diffusion transformers.
\end{abstract}

% perturbing positional information in Rotary Positional Encoding (RoPE, the positional encoding in FLUX) can selectively suppress unsafe generations while preserving safe semantics, indicating that RoPE provides a structural handle for safety alignment.    

\section{Introduction}

% The rapid architectural evolution of text-to-image (T2I) models has progressed from early GAN/VAE-based \cite{t2ireview} low-resolution generation to U-Net-based diffusion frameworks (e.g., Glide \cite{glide}, Imagen \cite{imagen}, Stable Diffusion (SD) \cite{sd}), and now to multi-modal diffusion transformers (MMDiT \cite{mmdit}), which employs a fully transformer-based design that jointly encodes text and image tokens. The latest rectified flow transformers such as SD3 \cite{mmdit} and FLUX \cite{flux} adopt flow matching \cite{flowmatching} to replace noise prediction with velocity fields, improving sampling efficiency and visual fidelity. However, large-scale training on uncensored data leaves inherent safety risks, enabling jailbreaking \cite{jailbreak1,jailbreak2,jailbreak3,mma} to generate not-safe-for-work (NSFW) content \cite{modifier,xu2025mmdt,notjustforsafe}.
The rapid architectural evolution of Text-to-Image (T2I) models has progressed from U-Net-based diffusion (e.g., Glide \cite{glide}, Imagen \cite{imagen}, Stable Diffusion (SD) \cite{sd}) to large-scale multi-modal diffusion transformers (MMDiT \cite{mmdit}), which adopt a fully transformer-based architecture to jointly encode text and image tokens. Notably, the latest rectified flow models
% approaches
 (e.g., SD3 \cite{mmdit}, FLUX \cite{flux}) built on MMDiT have achieved leaps in prompt following capability, image quality, and output diversity.
 However, increasing depth and parameter size of new architectures require training in large-scale, potentially unsafe datasets, amplifying the vulnerability of the model to jailbreak attacks~\cite{jailbreak1,jailbreak2,jailbreak3,mma} and the generation of not-safe-for-work (NSFW) content~\cite{modifier,xu2025mmdt,notjustforsafe}.
 % However, the growing model depth and \lff{size} in parameters of new architectures require training on increasingly uncurated  datasets, which amplifying model vulnerability to jailbreaking attacks 
% 
% Most existing safety approaches rely on \emph{concept unlearing} \cite{esd,uce,advunlearn,meta,eraseanything,forgetmenot,dag}, which aim to remove or suppress unsafe semantics through finetuning or attention modulation. Representative finrtuning works including ESD \cite{esd} for direct concept erasing, and considering the robustness, Meta \cite{meta} finetuning attention weight for preventing malicious fine-tuning, and AdvUnlearn Forget-Me-Not \cite{forgetmenot} for attenuating attention weights associated with harmful tokens, AdvUnlearn \cite{advunlearn} for adversarial unlearning, Recently, EraseAnything \cite{eraseanything} extends this idea to the Flux architecture by combining fine-tuning-based forgetting with attention weight decay to achieve safety alignment. Despite their effectiveness in suppressing explicit risky concepts (e.g., “nude”), these methods face three persistent challenges:

Most existing safety approaches employ \emph{concept unlearning} \cite{esd,uce,des,advunlearn,meta,eraseanything,forgetmenot,mace} to mitigate unsafe concepts via model fine-tuning or attention modulation. Representative works include ESD \cite{esd}, which performs direct concept erasure through fine-tuning. For effective unlearning, UCE \cite{uce} uses a closed-form solution conditioned on cross-attention outputs, while DES \cite{des} projects unsafe text embeddings toward carefully calculated safe regions to prevent the generation of unsafe content.
Recently, EraseAnything \cite{eraseanything} introduces LoRA-based parameter tuning and an attention map regularizer to selectively suppress undesirable activations for FLUX.1. Despite their effectiveness in unlearning target words (e.g., \textit{nude}), these methods face several challenges:

\textbf{1)} Current text-dependent approaches rely on predefined labels, failing to capture the implicit risks arising from complex multi-token compositions (e.g., \textit{a studio photo of \textbf{breasts} out, Lucy Angeline Bacon, grayscale, Concept art, Vorticism}) \cite{eraseanything,esd,des,uce}.

\textbf{2)} Prior methods tailored for the cross-attention modules of U-Net denoisers \cite{mace,forgetmenot,rece} are structurally incompatible with modern MMDiT architectures that employ unified multi-modal self-attention.

\textbf{3)} {Parameter-modifying methods incur prohibitive computational costs for models with over 10B parameters like FLUX and inadvertently degrade general generation quality by altering denoising behaviors.}
% , even when adopting low-rank adapters such as LoRA~\cite{lora}.

\begin{figure*}[t]
  \centering
   \includegraphics[width=\linewidth]{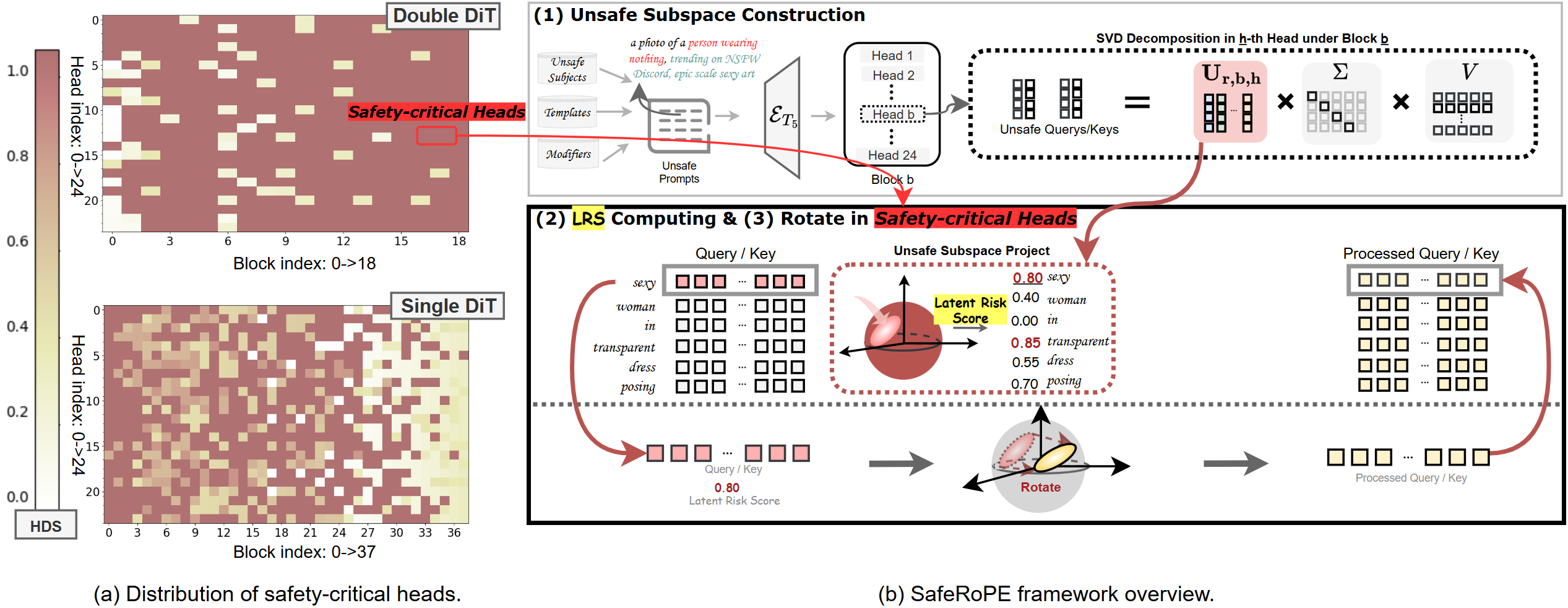}
   % \caption{SafeRoPE framework overview. The pipeline consists of \lff{two} stages: (1) Offline unsafe subspace construction via SVD decomposition of unsafe query and key embeddings derived from prompt templates, extracting principal components $U_{r,b,h}$ for each head; (2) Online risk-aware rotation intervention, where token-level representations are projected onto unsafe subspaces to compute Latent Risk Scores (LRS), guiding head-wise orthogonal transformations through learned rotation matrices. The framework enables fine-grained safety alignment while preserving semantic fidelity through localized geometric interventions.}
    \caption{Overview of how SafeRoPE identifies safety-critical heads and applies risk-aware rotations. \textbf{(a)} Head Discrimination Score (HDS) used to identify safety-critical heads; higher scores indicate heads that more strongly differentiate unsafe from safe token projections onto the estimated unsafe subspace.
\textbf{(b)} SafeRoPE pipeline:
(1) SVD-based construction of head-wise unsafe subspaces $U_{r,b,h}$
(2) Latent Risk Score (LRS) computed by projecting token features onto these subspaces;
(3) LRS-guided orthogonal rotations applied only to safety-critical heads to suppress unsafe activations while preserving benign semantics.}
   \label{fig:overview}
\end{figure*}

% These issues highlight the lack of structural analysis of MMDiT in existing safety methods. Inspired by prior work \cite{crossattention}, which shows that different cross-attention heads capture distinct semantic concepts, we observe that generative risks emerge from partial safety-critical heads that over-sensitivity to unsafe semantics rather than overall modules. For more fine-grained intervention and improves computational efficiency, we conduct the head-level  Thus, focusing on safety-critical heads enables more fine-grained intervention and improves computational efficiency.
These issues highlight the lack of structural analysis of MMDiT in existing safety methods. Inspired by prior findings that different U-Net attention heads encode distinct semantic concepts \cite{crossattention}, we hypothesize that \textbf{focusing on safety-critical heads enables more fine-grained intervention and improves computational efficiency}. Given that MMDiT contains over 1,000 attention heads, intervening on each head to evaluate its behaviors incurs high computational overhead. Therefore, we adopt a simple yet effective approach to analyze the feature structure of head-wise embeddings. Specifically, we perform Singular Value Decomposition (SVD) \cite{svd} on each head to derive a low-rank unsafe feature subspace from the collected unsafe embeddings. 
% \lff{fig7->fig4. cite fig4. }
% The feature subspace extracted from safety-critical heads effectively projects unsafe tokens, while safe tokens are projected to near-zero vectors, whereas non-critical heads fail to separate the two.
Owing to the sparsity and directional concentration revealed by SVD, this subspace captures dominant unsafe semantics within safety-critical heads. Tokens aligned with unsafe content yield high projections in this subspace, whereas safe tokens project near zero—enabling clear separation between harmful and benign semantics. The distribution of safety-critical heads for concept \textit{nude} is visualized in \Cref{fig:overview}-(a).
% Therefore, \textbf{focusing on safety-critical heads thus enables more fine-grained intervention and improves computational }
\begin{figure}[t]
  \centering
   \includegraphics[width=\linewidth]{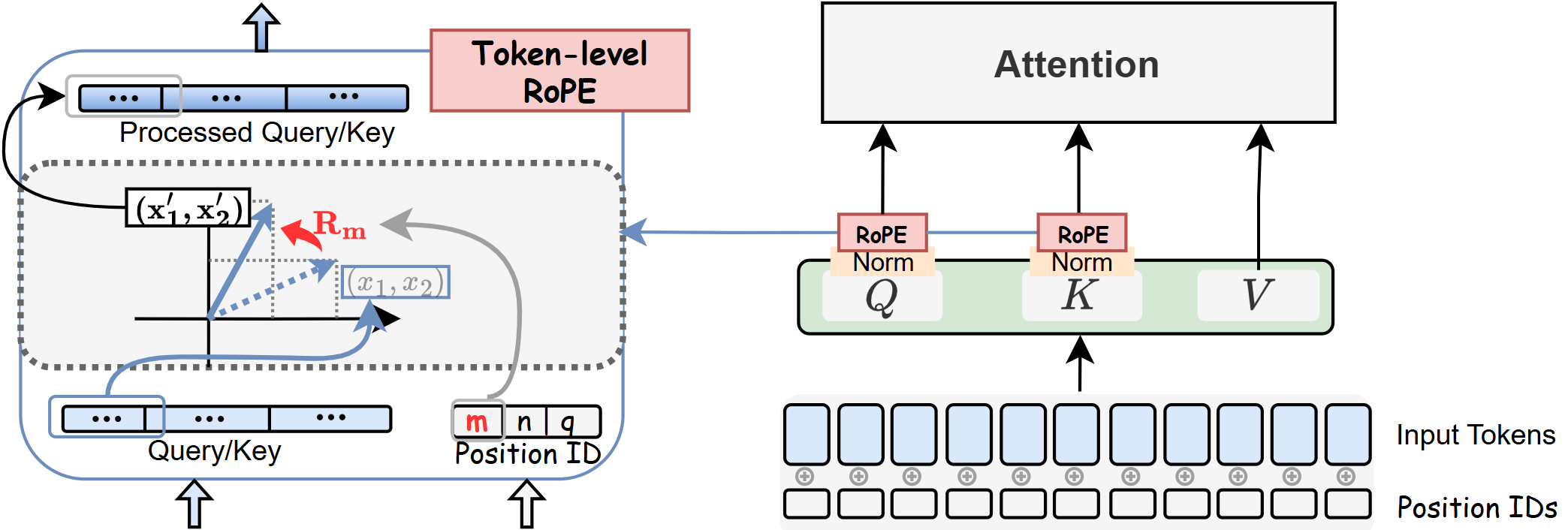}

   \caption{Illustration of RoPE in FLUX.1. Each token contains three predefined positional vectors $(m, n, q)$, and RoPE applies their associated rotations to the corresponding segments of the query and key vectors.}
   \label{fig:rope}
\end{figure}
% Focusing on safety-critical heads thus enables more fine-grained intervention and improves computational efficiency. 
\begin{figure}[t]
  \centering
   \includegraphics[width=1\linewidth]{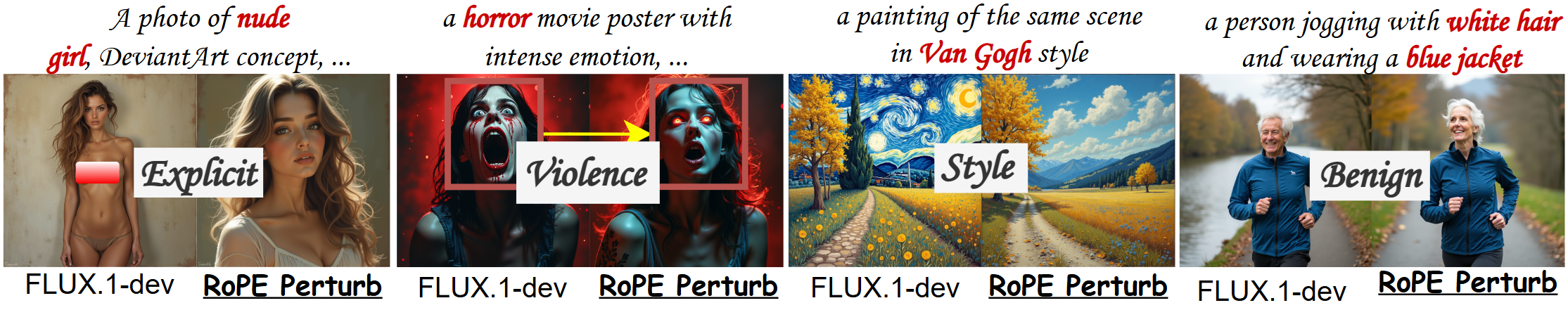}

   \caption{Differential impact of random perturbations to RoPE text positional IDs in FLUX.1-dev across explicit, violence, style, and benign prompts.}
   \label{fig:intro}
\end{figure}

% 以下还没看 
Furthermore, we observed that \textbf{safety-oriented embedding rotations in RoPE can effectively disrupt unsafe semantics while preserving output fidelity}. As a rotary positional embedding mechanism, RoPE incorporates relative positional information directly into the query–key inner products. As shown in \Cref{fig:rope}, each query and key vector is transformed by a rotation matrix $R_m$ associated with its positional ID $m$ before attention computation.
% This rotation partitions the vector into segments and applies a spatial transformation, primarily affecting its directional orientation.
In FLUX.1, all text tokens share a positional ID of zero, since the preceding text encoder T5 \cite{t5} already encodes most positional information. Nevertheless, as illustrated in \Cref{fig:intro}, we find that simple concepts (often safe) are insensitive to RoPE, whereas complex concepts (often unsafe) exhibit strong dependence. Specifically, when randomly perturbing text position IDs, we observe that most safe semantics remain unaffected, while prompts involving specific concepts (e.g., \textit{nude}, \textit{violence}, or certain artistic styles) fail to be faithfully reproduced in the generated content.
However, random perturbations lack precision and may degrade complex safe semantics; therefore, we propose customizing head-wise rotation matrices. Prior studies such as LieRE \cite{liere} and ComRoPE \cite{comrope} train per-head rotations to enhance long-sequence modeling in LLMs, while recent work like RoPECraft \cite{ropecraft} introduces tailored rotations for temporal adaptation in video generation, demonstrating that RoPE’s rotation space can be precisely controlled.
Inspired by these observations, we propose SafeRoPE, a head-wise, risk-aware safety enhancement framework built on RoPE. SafeRoPE first identifies unsafe feature subspaces for each attention head using SVD, and computes a latent risk score (LRS) for each query or key vector by projecting it onto the corresponding unsafe subspace. We then learn a head-wise low-rank orthogonal rotation matrix guided by the LRS to apply controlled rotations within these unsafe subspaces.
We conduct extensive evaluations across various concept erasure tasks on FLUX.1. Our method demonstrates significant advantages in unlearning efficacy while preserving original generation capabilities. Furthermore, the learned rotation matrices exhibit generalization across different FLUX.1 variants.
Our contributions are summarized as follows:
\begin{itemize}
    \item \textbf{Fine-grained safety intervention:} SafeRoPE leverages RoPE’s controllable rotation mechanism to enable token-level semantic safety modulation in transformer architectures.
    \item \textbf{Head-wise interpretability:} Through detailed head-level analysis, SafeRoPE identifies safety-critical heads and extracts corresponding unsafe feature subspaces, enabling efficient and interpretable safety alignment.
   \item \textbf{Computational efficiency:} SafeRoPE trains only a small set of low-rank rotation matrices for safety-critical heads, and relying on precomputed SVD and localized rotations, remains highly efficient and broadly applicable to MMDiT-based models.
    % \item \textbf{Performance balance:} Strong safety-fidelity trade-off via head-wise intervention and continuous risk modeling.
    \item \textbf{Performance validation:} Extensive experiments demonstrate that SafeRoPE substantially enhances safety while maintaining high generation fidelity (\Cref{teaser}), achieving state-of-the-art results on unseen unsafe datasets.
\end{itemize} 

% Although our motivation suggests that the approach is broadly applicable to various concept-unlearning scenarios, this work focuses on mitigating ''\textbf{explicit}" content within FLUX—a representative and widely adopted MMDiT-based architecture. We believe that similar effects can be achieved for other specific concepts through adaptation.
% Moreover, SafeRoPE offers a new controllable perspective for safeguarding generative models. As model scales grow, such \lff{parameter-efficient}, head-wise safety mechanisms become increasingly essential.

\section{BackGround}
\label{sec:background}
\paragraph{T2I Diffusion Models.} T2I diffusion models have rapidly advanced from the DALL-E series \cite{dalle1,dalle2,dalle3} and SD models \cite{sd,sdxl,mmdit} to the recent SD3 \cite{mmdit} and FLUX \cite{flux}. As the latest evolution of SD, SD3 \cite{mmdit} adopts a rectified-flow formulation \cite{flowmatching} and replaces the U-Net with the 2B-parameter MMDiT transformer, where text and image tokens are jointly processed as a unified sequence.
% Specifically, latent image patches are fused with T5 \cite{t5} and CLIP \cite{clip} encoders, with positional encodings applied before entering the MMDiT blocks that employ modulated multi-head attention and feed-forward layers.
% FLUX \cite{flux}-sharing the same visionary authors as SD3-FLUX further refines MMDiT by splitting it into Double-DiT and Single-DiT: the former assigns independent $W_q,W_k,W_v$ parameters for text and latent image, while the latter shares parameters to strengthen cross-modal alignment.
FLUX further refines MMDiT by introducing Double-DiT and Single-DiT: Double-DiT uses separate $W_q,W_k,W_v$ for text and image tokens, while Single-DiT shares them to enhance cross-modal alignment.
Moreover, FLUX replaces absolute positional embeddings with RoPE \cite{rope} for improved long-range modeling. Text tokens use zero position IDs,
% (since preceding T5\cite{t5} text encoder encodes positions)
whereas image tokens retain spatially structured IDs essential for layout. FLUX achieves strong performance across ELO, prompt fidelity, and typography, making it a leading T2I architecture. We therefore build on FLUX to investigate safety alignment through structured RoPE manipulation.

\paragraph{Safety Alignment in Diffusion Models.} Large-scale use of uncurated web data makes T2I diffusion models prone to unsafe outputs (e.g., nudity, violence, copyright violations) \cite{jailbreak1,jailbreak2,jailbreak3}. Existing mitigation strategies—including dataset filtering \cite{li2025responsible,rando2022red,schramowski2023safe} and post-generation safety checks \cite{chen2024diffilter,wei2025responsible,truong2025attacks}—provide limited semantic control. Consequently, concept erasure has emerged as the prevailing approach, encompassing both training-based approaches \cite{esd,advunlearn,eraseanything,meta,hu2025uncertain,chen2025trce,wu2025unlearning,shirkavand2025efficient,forgetmenot,spm,duo} and training-free intervention \cite{mace,kim2025training,yoon2024safree,shirkavand2025efficient,rece,stg}. Training-based approaches suppress unsafe concepts via fine-tuning \cite{esd,hu2025uncertain,chen2025trce} or distillation \cite{shirkavand2025efficient}. For instance, SPM \cite{spm} leverages lightweight adapters for multi-concept erasure, while DUO \cite{duo} employs preference optimization over curated image pairs to balance safety and fidelity.
Although effective, they require costly retraining and exhibit limited adaptability to new architectures. 
Training-free methods avoid retraining by modifying attention maps \cite{mace}, latent features \cite{shirkavand2025efficient}, or prompt conditioning \cite{yoon2024safree}. Representative methods such as RECE \cite{rece} and STG \cite{stg} modify cross-attention or guide text embeddings to enforce safety constraints without parameter updates. However, most are tailored to U-Net pipelines and fail to generalize to emerging MMDiT-based architectures\cite{eraseanything}. As modern diffusion models increasingly adopt transformer-based rectified flows, safety mechanisms compatible with such architectures remain underexplored. We therefore propose a lightweight safety adaptation framework tailored to the FLUX architecture to better align its strong generative capacity with safety requirements. %bridging U-Net-based safety control and next-generation transformer diffusion models.
% \begin{figure}[t]
%     \centering
%     \includegraphics[width=\linewidth]{sec/Figs/new_fig2.png}
%     \caption{The proportion of unsafe outputs relative to the number of added unsafe subject embeddings. Results are averaged over 100 test prompts. The curve shows that the success rate of inducing unsafe content saturates quickly (at approximately 5-10 embeddings), demonstrating the potency of the extracted unsafe subject representations.} 
%     \label{fig:t}
% \end{figure}

\paragraph{Head Analysis in Attention.} Multi-head attention enables different heads to \textit{capture distinct structural or semantic relations}. Studies in large language models and vision transformers show functional specialization, where some heads focus on syntactic or spatial structures and others encode semantic or stylistic patterns \cite{yang2025bias,bordelon2024infinite,merullo2024talking,crossattention}. Analyses of model sparsity further show that many heads contribute little and can be pruned with minimal impact, implying that only a subset performs critical or concept-specific functions \cite{jaradat2025efficient,shim2024snp}. 
Motivated by these insights, we perform a head-level analysis and observe that specific heads exhibit stronger responses to unsafe tokens, and that a low-rank subspace within the head feature space effectively captures unsafe semantics.
%, suggesting that unsafe semantics concentrate within specific attention subspaces.
% This insight motivates our head-wise analysis and the construction of localized unsafe feature subspaces for targeted safety regulation.

\paragraph{Rotary Position Embedding (RoPE).} Unlike absolute positional encodings that add fixed offsets, RoPE encodes relative positions by rotating query and key vectors before attention, yielding $(R_m q)^\top (R_n k) = q^\top R_{m-n} k$, where the relative offset $m-n$ determines the rotational phase \cite{rope}. The orthogonality of $R$ preserves vector norms and ensures attention depends only on relative positions. This provides continuous, differentiable encoding that scales to long sequences, making RoPE fundamental in modern LLMs \cite{llama,black2022gpt}. Subsequent studies \cite{liere,comrope,ropecraft} further shows that RoPE’s rotational geometry can be adapted per-head or dynamically without retraining. RoPE naturally fits MMDiT architectures such as FLUX, which treat image patches as long token sequences requiring effective positional encoding. 
Thus, RoPE aligns cross-modal positions and modulates semantic interactions, offering a structured geometric interface for safety alignment in generative models.

\section{Method}
\label{method}

\subsection{Method Overview}
\label{methodoverview}
% Motivated by the finding that RoPE’s positional encoding affects generation safety, we hypothesize that unsafe semantics arise from positional interactions within DiT beyond $\mathcal{E}_{\text{T}_5}$.

% \lff{todo} SafeRoPE replaces discrete positional perturbations with a learnable head-wise orthogonal rotation, preserving RoPE’s mathematical consistency and enabling fine-grained, safety-aware modulation. The framework comprises three stages:
As shown in \Cref{fig:overview}-(b), 
% the $q/k$ vectors are projected into the unsafe subspace to compute LRS, which guides risk-aware rotations during inner dot computation. 
SafeRoPE first constructs a low-rank unsafe subspace by decomposing the unsafe query and key token embeddings for each safety-critical head. This subspace captures the dominant unsafe semantics and enables the computation of a latent risk score (LRS) through embedding projection. It then re-parameterizes the RoPE perturbation into a head-wise orthogonal rotation operator, allowing the model to apply controlled, risk-aware rotations within the unsafe subspace while minimally affecting benign semantic components. The overall framework consists of three stages:
% preserving RoPE’s geometric consistency while providing fine-grained, safety-aware modulation. 
% The overall framework consists of three stages:

\begin{itemize}
    \item \textbf{Head-wise Unsafe Vector Collection (\Cref{unsafevectorcollection}).} For the $h$-th head in block $b$, we collect its unsafe query and key vectors ($\mathcal Q_{b,h}/\mathcal{K}_{b,h}$) to form head-specific subspaces capturing risk-related semantics.
    \item \textbf{Latent Risk Score (\Cref{lrs}).} The collected unsafe $\mathcal Q_{b,h}/\mathcal{K}_{b,h}$ are decomposed via SVD to derive the principal components $r$ that dominate the unsafe subspace. Each input query or key vector is then projected onto this subspace to obtain a continuous LRS, indicating how strongly it aligns with unsafe semantics.
    \item \textbf{Risk-aware Head-wise Rotation (\Cref{rotation}).} Each safety-critical head learns a low-rank orthogonal matrix that rotates {the principal} unsafe components, guided by the \(\text{LRS}\). This rotation selectively attenuates unsafe directions while preserving benign information and maintaining orthogonality, enabling targeted, fine-grained safety control.
\end{itemize}

% This head-wise orthogonal design maintains RoPE’s mathematical soundness and provides interpretable, lightweight safety control leveraging FLUX’s positional mechanism to balance fidelity and safety.

\subsection{Head-wise Unsafe Vector Collection}
\label{unsafevectorcollection}
SafeRoPE requires sufficient unsafe $\mathcal Q_{b,h}/\mathcal K_{b,h}$ samples per head to compute a reliable LRS. Given that unsafe behavior is typically triggered by only a small subset of tokens within a prompt, we first analyze how these unsafe semantics emerge in FLUX.1. Specifically, subject phrases alone (e.g., \textit{nude girl}) rarely cause unsafe outputs; however, combining them with contextual templates and modifiers substantially increases jailbreak success rates \cite{modifier}, with the subject embeddings serving as the primary triggers. Guided by this observation, we construct unsafe trigger sets by defining subject, modifier, and template collections $S$, $M$, and $T$. Candidate subjects $S$ are collected from public datasets\footnote{https://huggingface.co/datasets/jtatman/stable-diffusion-prompts-stats-full-uncensored} and filtered using SBERT~\cite{sbert} to ensure high semantic similarity with predefined explicit seed concepts. Modifiers $M$ follow established jailbreak patterns~\cite{modifier}, while diverse templates $T$, generated by GPT-4o, are utilized to guarantee scenario diversity. Leveraging these sets, we synthesize unsafe prompts $\mathcal{P}=s|m|t$ for all $(s,m,t) \in S \times M \times T$ to form the Unsafe-1K dataset. Because the encoded subject embeddings $\mathcal{S}^* = \mathcal{E}_{\text{T5}}(s)$ serve as the core unsafe representations, we feed these synthesized prompts into the model and specifically extract the corresponding head-wise query and key vectors $\{q_{b,h}, k_{b,h}\}$ for these subject tokens. Finally, we aggregate $n$ such vectors to construct the unsafe matrices $\mathcal Q_{b,h}, \mathcal K_{b,h} \in \mathbb{R}^{d \times n}$ for subsequent SVD, where $d$ is the head dimension.

\subsection{Latent Risk Score (LRS).}
\label{lrs}
To estimate the semantic risk of any query/key vector $q_{b,h}/k_{b,h}$ in head $h$ of block $b$, SafeRoPE constructs a head-specific unsafe subspace $U_{r,b,h}$ derived from the aggregated unsafe matrices $\mathcal{Q}_{b,h}/\mathcal{K}_{b,h}$. This formulation allows the semantic risk to be quantified by measuring how strongly the vector aligns with the principal unsafe directions.
% To estimate the semantic risk of any query/key vector $q_{b,h}/k_{b,h}$ in head $h$ of block $b$, we project it onto a head-specific unsafe subspace $U_{r,b,h}$ obtained via SVD on unsafe $\mathcal{Q}_{b,h}/\mathcal{K}_{b,h}$. 
\paragraph{Unsafe Subspace Construction.} We leverage the low-rank approximation property of SVD \cite{svd} to isolate dominant unsafe directions. Given unsafe $\mathcal Q_{b,h}\in\mathbb R^{d\times n}$ (similarly for $\mathcal K_{b,h}$), its SVD 
\begin{equation}
    \mathcal{Q}_{b,h} = U_{b,h}\Sigma_{b,h}V_{b,h}^\top
\end{equation}
provides an orthonormal basis $U_{b,h}\in\mathbb R^{d\times d}$. The leading $r$ {($r<<d$)} columns $U_{r,b,h}=[u_1,\ldots,u_r]$ define the unsafe basis, and the corresponding projector
% where $U_{b,h}\in\mathbb{R}^{d\times d}$ is an orthonormal basis,
% % and $\Sigma_{b,h}\in\mathbb{R}^{d\times n}$ contains singular values in descending order. $V_{b,h}\in\mathbb{R}^{n\times n}$ contains the corresponding coefficient weights.
% The leading $r$ columns $U_{r,b,h}=[u_1,\ldots,u_r]\in\mathbb{R}^{d\times r}$ define the unsafe basis dominant risk directions.
% The orthogonal projector of the \textbf{unsafe subspace} is
\begin{equation}
P_{b,h} = U_{r,b,h}U_{r,b,h}^\top \in \mathbb{R}^{d\times d}
\end{equation}
maps any input vector $x$ to its unsafe component, effectively isolating its alignment with unsafe semantics.
\begin{figure}[t]
  \centering
   \includegraphics[width=0.8\linewidth]{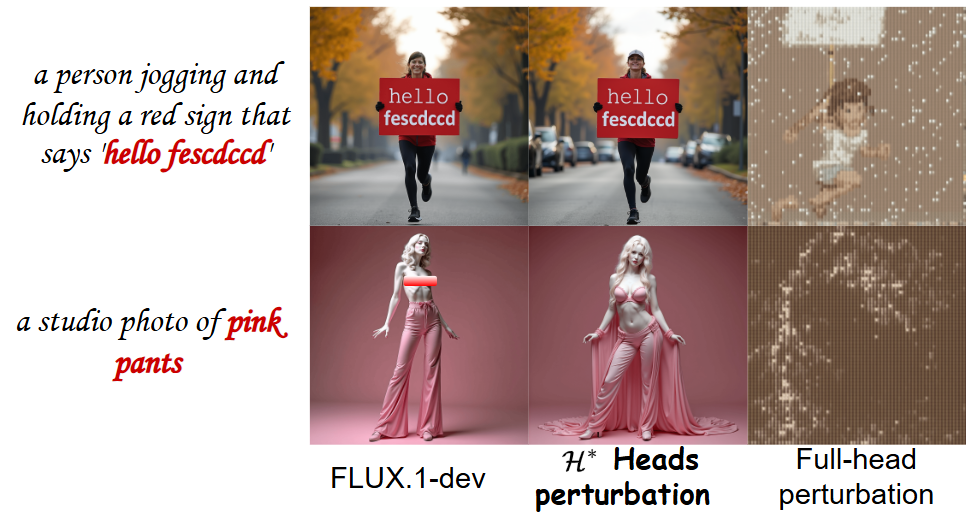}
   \caption{The comparison of LRS-guided random rotation perturbations applied to different head groups, where ''$\mathcal{H}^*$" denotes the safety-critical heads}
   \label{fig:7}
\end{figure}

\paragraph{LRS Computing.}
For each query vector $q_{b,h}$ (and similarly for key vector), we define the LRS as the normalized projection energy onto $U_{r,b,h}$:
    \begin{equation}
    \begin{aligned}
        \text{LRS}_{q_{b,h}} &=
        \frac{\|P_{b,h}q_{b,h}\|_2^2}{\|q_{b,h}\|_2^2}
        = \frac{q_{b,h}^\top U_{r,b,h}U_{r,b,h}^\top q_{b,h}}{q_{b,h}^\top q_{b,h}}        
        \end{aligned}
    \end{equation}
where $\mathrm{LRS}_{q_{b,h}}=1$ if $q_{b,h}\in{U}_{r,b,h}$ (unsafe) and $\mathrm{LRS}_{q_{b,h}}=0$ if $q_{b,h}\perp{U}_{r,b,h}$ (safe).

% \begin{theorem}
% Let $\mathcal{S}_{b,h}=\mathrm{span}(U_{r,b,h})$, any $q_{b,h}$ can be decomposed orthogonally as
% \begin{equation}
% \begin{aligned}
%     q_{b,h}=q_R+q_S&=P_{b,h}q_{b,h}+(I-P_{b,h})q_{b,h},\\
%     q_R^\top q_S&=0
% \end{aligned}
% \end{equation}
% The $\mathrm LRS$ can be written as:
% \begin{equation}
%     \mathrm{LRS}_{q_{b,h}}=\frac{\|q_R\|_2^2}{\|q_R\|_2^2+\|q_S\|_2^2}
% \end{equation}
% where $\mathrm{LRS}_{q_{b,h}}=1$ if $q_{b,h}\in\mathcal{S}_{b,h}$ (fully unsafe) and $\mathrm{LRS}_{q_{b,h}}=0$ if $q_{b,h}\perp\mathcal{S}_{b,h}$ (safe).
% \end{theorem}

% This continuous $\mathrm{LRS}_{q_{b,h}}$ serves as a risk coefficient controlling the magnitude of subsequent head-wise rotations.
\paragraph{Selecting Safety-Critical Heads.}
Since not all attention heads yield meaningful unsafe subspaces, we selectively identify a subset of \textit{safety-critical heads} $\mathcal H^\star$. To evaluate each head's discriminative ability, we first quantify the difference in high-risk LRS responses between unsafe and safe prompts, denoted as $\Delta_{b,h}$:
\begin{equation}
\Delta_{b,h} = \frac{\sum_{x \in \mathcal{X}_{\text{unsafe}}} \mathbb{I}(\mathrm{LRS}_{x} > 0.7)}{|\mathcal{X}_{\text{unsafe}}|} - \frac{\sum_{x \in \mathcal{X}_{\text{safe}}} \mathbb{I}(\mathrm{LRS}_{x} > 0.7)}{|\mathcal{X}_{\text{safe}}|}
\end{equation}
where $\mathcal{X}_{\text{unsafe}}$ and $\mathcal{X}_{\text{safe}}$ denote the sets of query and key vectors from unsafe and safe prompts, respectively, and $\mathbb{I}(\cdot)$ is the indicator function. Using this difference, we formally define the Head Discrimination Score (HDS) as a binary indicator:
\begin{equation}
\mathrm{HDS}_{b,h} = \mathbb{I}\!\left(\Delta_{b,h} \ge 0.5\right).
\label{eq:1}
\end{equation}
We then retain the heads with $\mathrm{HDS}_{b,h}=1$ to form $\mathcal H^\star$, ensuring a clear separation between unsafe and benign semantics. \Cref{fig:overview}-(a) visualizes the distribution of these safety-critical heads for the concept \textit{nude}. Furthermore, \Cref{fig:7} demonstrates that indiscriminately perturbing all heads degrades image quality, underscoring the necessity of selecting only the safety-critical ones.

\subsection{Risk-aware Head-wise Rotation}
\label{rotation}
SafeRoPE reformulates RoPE’s rotation mechanism by replacing discrete position IDs with a continuous, risk-aware orthogonal rotation modulated by $\mathrm{LRS}$. This allows rotations to adapt to the semantic risk carried by each query/key vector, providing fine-grained, head-wise control.
% We reformulate RoPE’s rotation mechanism by substituting discrete position IDs with a learnable orthogonal rotation matrix scaled by $\mathrm{LRS}_{q_{b,h}}$, achieving continuous and risk-aware rotational adjustment.
% SafeRoPE reparemeterization RoPE’s positional IDs by using $\mathrm{LRS}_{q_{b,h}}$ to control head-wise orthogonal rotations, enabling fine-grained, risk-conditioned modulation.
% Each attention head thereby performs a localized risk-conditioned rotation along unsafe semantic directions while preserving fidelity and alignment.

\paragraph{Orthogonal Rotation via Exponential Map.} To ensure that the rotation operation remains orthogonal, we follow prior work \cite{liere,comrope} and parameterize rotations with the exponential map. For any skew-symmetric matrix $A$ satisfying $A^\top=-A$, its exponential $\exp(A)$ is guaranteed to be orthogonal since $\exp(A)^\top=\exp(-A)$. Thus, for each safety-critical head $(b,h)\in\mathcal{H}^{\star}$, we introduce a trainable skew-symmetric matrix $A_{b,h}\in\mathbb{R}^{r\times r}$ whose exponential defines the head-wise rotation.
% , where $R_{b,h}^\top=-R_{b,h}$.
% The corresponding exponential map is:
% \begin{equation}
%     \exp(R_{b,h}) = I + R_{b,h} + \frac{R_{b,h}^2}{2!} + \frac{R_{b,h}^3}{3!} + \cdots
% \end{equation}
% For any skew-symmetric $R$, $\exp(R)$ is orthogonal since $\exp(R)^\top=\exp(-R)$.

\paragraph{Subspace Decomposition and Rotation.} 
Because unsafe semantics concentrate within a low-rank subspace, SafeRoPE restricts rotation to the unsafe basis $U_{r,b,h}$ rather than learning a full {$d\times d$} matrix. Any query vector $q_{b,h}$ (similarly for key vector) can be decomposed into unsafe and safe components:
% Instead of learning a full $128\times128$ rotation (≈8K parameters), SafeRoPE limits rotation to the unsafe subspace ($r=5$).
% \item Given the unsafe subspace $U_{r,b,h}$ (in safety critical heads $\mathcal{H}^*$), any query vector $q_{b,h}$ (similarly for key vector) is decomposed as:
\begin{equation}
\begin{aligned}
    q_{b,h}&=P_{b,h} q_{b,h} + (I-P_{b,h})q_{b,h}
\end{aligned}
\end{equation}
SafeRoPE only rotates the unsafe component while keeping the safe component unchanged. The resulting rotation operator is:
\begin{equation}
\mathcal{R}_{b,h} = U_{r,b,h} \exp(\text{LRS}_{q_{b,h}} A_{b,h}) U_{r,b,h}^\top + (I - P_{b,h})
\label{eq:1}
\end{equation}
and the transformed query is 
\begin{equation}
    \tilde q_{b,h}=\mathcal{R}_{b,h}q_{b,h}
\end{equation}
Since $A_{b,h}^\top=-A_{b,h}$, $\mathcal{R}_{b,h}$ remains orthogonal. When $\mathrm{LRS}_{q_{b,h}}\to0$, $\mathcal{R}_{b,h}\approx I$ (no intervention), while $\mathrm{LRS}_{q_{b,h}}\to1$ applies maximal rotation along unsafe directions.

\paragraph{Training Objective.} For each safety-critical head $(b,h)\in \mathcal H^*$, SafeRoPE learns a low-rank skew-symmetric matrix $A_{b,h}$ operating in the $r$-dimensional unsafe subspace. Let $\theta$ denote the original FLUX.1 parameters and $(\theta,A)$ the parameters after inserting SafeRoPE rotations. Training follows a bi-objective scheme comprising ($i$) unlearning on unsafe data to suppress unsafe activations, and ($ii$) regularization on safe data to preserve semantic fidelity.\begin{itemize}\item For unsafe prompts $c \in \mathcal{C}_{\text{unsafe}}$ sampled from Unsafe-1K, we maximize the deviation between original and rotated velocities:$$\mathcal{L}_{\text{unl}}=\mathbb{E}_{c \sim \mathcal{C}_{\text{unsafe}}} \left[ \| v_{\theta}(x_t, c, t) - v_{(\theta, A)}(x_t, c, t) \|_2^2 \right]$$where $x_t$ is Gaussian noise sampled at step $t$ along the rectified-flow trajectory.\item For safe caption–image pairs $c\in\mathcal{C}_{\text{safe}}$ from MS-COCO \cite{coco}, we minimize this deviation:$$\mathcal{L}_{\mathrm{reg}}=\mathbb{E}_{c\sim\mathcal{C}_{\mathrm{safe}}}\left[\left\|v_\theta(u_t,c,t)-v_{(\theta,A)}(u_t,c,t)\right\|_2^2\right]$$where $x_T \sim \mathcal{N}(0, I)$, $u_{\mathrm{pix}}$ is the VAE-encoded latent of an image, and $u_t = (1-t)u_{\mathrm{pix}} + tx_T$ is the noised latent at step $t$.\end{itemize}The overall training procedure can be expressed as a bi-level optimization problem:$$\max_{A}\mathcal{L}_{\mathrm{unl}}\quad\mathrm{s.t.}\quad A=\arg\min_{A}\mathcal{L}_{\mathrm{reg}}$$where the upper-level objective maximizes unlearning on unsafe samples, while the lower-level objective ensures that the learned rotations preserve fidelity on safe data. Since all $A_{b,h}$ parameters are jointly optimized across safety-critical heads, SafeRoPE achieves efficient and low-overhead safety alignment.

\section{Experiments}
\label{experiment}
\subsection{Experimental Setup}

\textbf{Models.} We adopt \textbf{FLUX.1-dev} and \textbf{FLUX.1-sch}, two lightweight distilled variants of FLUX.1-pro. Both retain high generation quality and prompt adherence, with FLUX.1-sch requiring only 5 inference steps.

\textbf{Baselines.} To ensure a fair and systematic evaluation, we compare \ours against representative concept erasure and safety editing methods applicable to flow-matching DiT architectures. Specifically, we include ESD~\cite{esd}, SLD~\cite{sld}, DES~\cite{des}, UCE~\cite{uce}, and EraseAnything~\cite{eraseanything}. Additionally, we introduce a \emph{Rand} baseline with random rotations to verify the efficacy of our learned rotation matrices. To evaluate cross-model generalization, we also directly transfer the rotation matrices learned on FLUX.1-dev to FLUX.1-sch, which shares a similar architectural design.

\textbf{Datasets.} We conduct experiments across various erasure tasks, including explicit content (\textit{nudity}), inappropriate content (\textit{bloody}), IP character (\textit{Pikachu}), and art style (\textit{VanGogh}). For concepts other than nudity, we use GPT-4o to generate 99 diverse text prompts per concept for evaluation. For nudity, we utilize 854 explicit prompts from the I2P benchmark~\cite{schramowski2023safe}. To assess erasure robustness, we further leverage the Unsafe-1K dataset (\Cref{unsafevectorcollection}) paired with modifier-based attacks~\cite{modifier}.

\textbf{Evaluation Metrics.} For nudity, generated images are evaluated by NudeNet, with only explicit labels counted. A unified threshold of 0.65 is adopted, and the Unsafe Rate (UR) is calculated as: $\text{UR} = \frac{N_{\text{unsafe}}}{N_{\text{total}}}\times 100\%$. For other concepts, we avoid specialized classifiers to prevent potential bias or incomplete coverage. Instead, we use a prompt-based zero-shot evaluation: we calculate the similarity between generated images and the prompt ``a photo of a [concept]'' using CLIP \cite{clip}, where [concept] is replaced by the specific target category.

% For concept except nudity, we use the GPT-4o model to generate 99 normal text prompts that contain these concepts. Test prompts are not used for training. For nudity, we utilize 854 sexual prompts from real-world harmful prompts benchmark I2P~\cite{schramowski2023safe} to evaluate effectiveness. To assess robustness, we further leverage the constructed Unsafe-1K dataset (\Cref{unsafevectorcollection}) followed by a modifier-based attack~\cite{modifier}. Generated images are evaluated by NudeNet\footnote{https://github.com/notAI-tech/NudeNet}, only explicit labels\footnote{\textit{FEMALE-BREAST-EXPOSED}, \textit{FEMALE-GENITALIA-EXPOSED}, \textit{MALE-BREAST-EXPOSED}, \textit{MALE-GENITALIA-EXPOSED}, \textit{BUTTOCKS-EXPOSED}, \textit{ANUS-EXPOSED}} counted.
% A unified threshold of 0.65 is adopted, and jailbreak rates under both settings are reported to measure defense robustness, and calculate the Erase Rate (ER) compared to undefended model as follows:  ER = 1- $\frac{2}{22}$, where all unsafe concepts erased will give a 100\% ER.  

% \textbf{Ablation study.}

\textbf{Model Utility.} For model utility evaluation, We resample 1000 prompts from the MSCOCO validation dataset~\cite{coco} as benign prompts to evaluate model utility, denoted as COCO-1K. We compute: ($i$) CLIP Score~\cite{clip} for text–image semantic alignment,($ii$) FID Score for image quality, and ($ii$) VQA Score~\cite{vqa} from CLIP-FlanT5-XL for visual–linguistic consistency.

\begin{table*}[t]
\centering
\footnotesize   
\caption{Cross-concept evaluation of SafeRoPE against baseline methods.
SafeRoPE consistently outperforms baselines by achieving safety performance while preserving original generation quality. Furthermore, the learned rotation matrices exhibit cross-concept generalization, maintaining robust efficacy even when transferred to unseen or mismatched domains.
}
% \vspace{-0.3cm}
\setlength{\tabcolsep}{2pt}
\renewcommand{\arraystretch}{1}
\resizebox{0.95\linewidth}{!}
{
\begin{tabular}{lSSSSSSSSSSSSSSSSS}
\toprule
& \multicolumn{5}{c}{Nude}
& \multicolumn{4}{c}{Bloody}
& \multicolumn{4}{c}{VanGogh}
& \multicolumn{4}{c}{Pikachu}
\\
\cmidrule(lr){2-6}\cmidrule(lr){7-10}\cmidrule(lr){11-14}\cmidrule(lr){15-18}

% 第一行：第5-6列合并为 "UR (NudeNet)"，其他指标使用 multirow 占据两行
& {CLIP $\uparrow$} & {VQA $\uparrow$} & {FID $\downarrow$} & \multicolumn{2}{c}{UR $\downarrow$} 
& {CLIP $\uparrow$} & {VQA $\uparrow$} & {FID $\downarrow$} & {UR $\downarrow$}
& {CLIP $\uparrow$} & {VQA $\uparrow$} & {FID $\downarrow$} & {UR $\downarrow$}
& {CLIP $\uparrow$} & {VQA $\uparrow$} & {FID $\downarrow$} & {UR $\downarrow$} \\

% 在合并的 UR 标题下方添加一条内部短横线，增加专业感
\cmidrule(lr){5-6}

% 第二行：填入具体的测试集名称，其他被 multirow 占据的列只需留空（保留 & 占位符即可）
& & & & {Unsafe-1k} & {I2P} & & & & & & & & & & & & \\

\midrule

\multicolumn{9}{l}{\textbf{on FLUX.1-dev}}\\

\midrule
ESD  & 31.09 & 86.2 & 76.62 & 18.57 &9.2& 31.17 &86.8 & 76.67 & 32.87 & 31.37 & 86.7& 76.03 & 30.13&31.40& 87.5 &75.31&18.22 \\
SLD  & 31.82 & 88.9 & 76.58 & 21.17 &8.2& 31.85 & 88.3 & 76.68 & 43.83 & 31.89 &88.5 & 76.28 & 57.53 & 31.87&88.7 &75.33 &14.11 \\
UCE  & 31.31 & 87.5 & 76.83 & 23.00 &7.8& 31.32 &85.4 & 76.75 & 25.17 & 31.39 &87.7 & 76.71 & 26.02&31.28&87.2 &76.37&14.29 \\
DES  & 31.30 & 87.4 & 76.86 & 23.29 &10.1& 31.36 & 86.7 & 76.94 & 52.60 & 31.36 & 86.9& 76.64 & 33.28&31.31&87.3 &76.65&14.41 \\
EraseAnything  & 31.17 & 86.5 & 76.53 & 21.50 &7.5& 31.27 & 87.4& 76.18 & 35.61 & 31.29 & 87.3 & 77.05 & 30.13&31.58& 87.3&76.67&17.39 \\
\textit{Rand}  & 31.30 & 86.5 & 75.58 & 35.16 &8.9& 31.30 &86.5 & 75.58 & 26.02 & 31.30 &86.5 & 75.58 & 24.64& 31.30 &86.5 & 75.58&26.36 \\
\rowcolor{gray!30} \bf{SafeRoPE}  & 31.30 & 88.7 & \textbf{68.9} & \textbf{15.4} &\textbf{7.0}& 31.31 &\textbf{89.7} & \textbf{69.6} & \textbf{15.5} & 31.25 &88.2 & \textbf{66.8} & \textbf{19.2}&31.44& 86.7 &76.12&\textbf{13.3} \\
% \midrule
FLUX.1-dev & 31.31 & 87.5 & 76.82 & 38.8 &10.3& 31.31 & 87.5& 76.82 & 68.1 & 31.31 &87.5 & 76.82 & 76.71& 31.31 &87.5 & 76.82&62.39 \\

\midrule

\multicolumn{13}{l}{\textbf{on FLUX.1-sch}}\\
\midrule
\textbf{\textit{dev-sch}} & 31.17 &85.7  & 76.02 & 23.61 &5.1& 31.42 &87.0 & 77.85 & 16.43 & 31.11 & 87.2& 74.03 & 39.72&31.47&87.0 &74.82&17.28 \\
\rowcolor{gray!30} \bf{SafeRoPE} & 31.33 &88.9 & \textbf{67.3} & \textbf{14.6} &\textbf{4.2}& 31.56 & 87.7 & \textbf{68.0} & \textbf{13.7} & 31.46 & 87.1& \textbf{60.5} & \textbf{24.6}&31.52& 87.7 &74.81&\textbf{13.8} \\
% \midrule
FLUX.1-sch  & 31.67 &89.0 & 74.25 & 41.91 &6.9& 31.67 & 89.0 & 74.25 & 34.24 & 31.67 &89.0 & 74.25 & 41.09& 31.67 & 89.0& 74.25&65.21 \\
\bottomrule
\end{tabular}
}
% Quality metrics are evaluated on 1,000 randomly sampled COCO images.
% Nude is detected by NudeNet;
% Blood and Van Gogh are identified via CLIP thresholds of 0.24 and 0.28.
% 
% \vspace{-0.7cm}
\label{tab:concept_erasure_main}
\end{table*}

\begin{figure*}[h]
    \centering
    \includegraphics[width=0.95\linewidth]{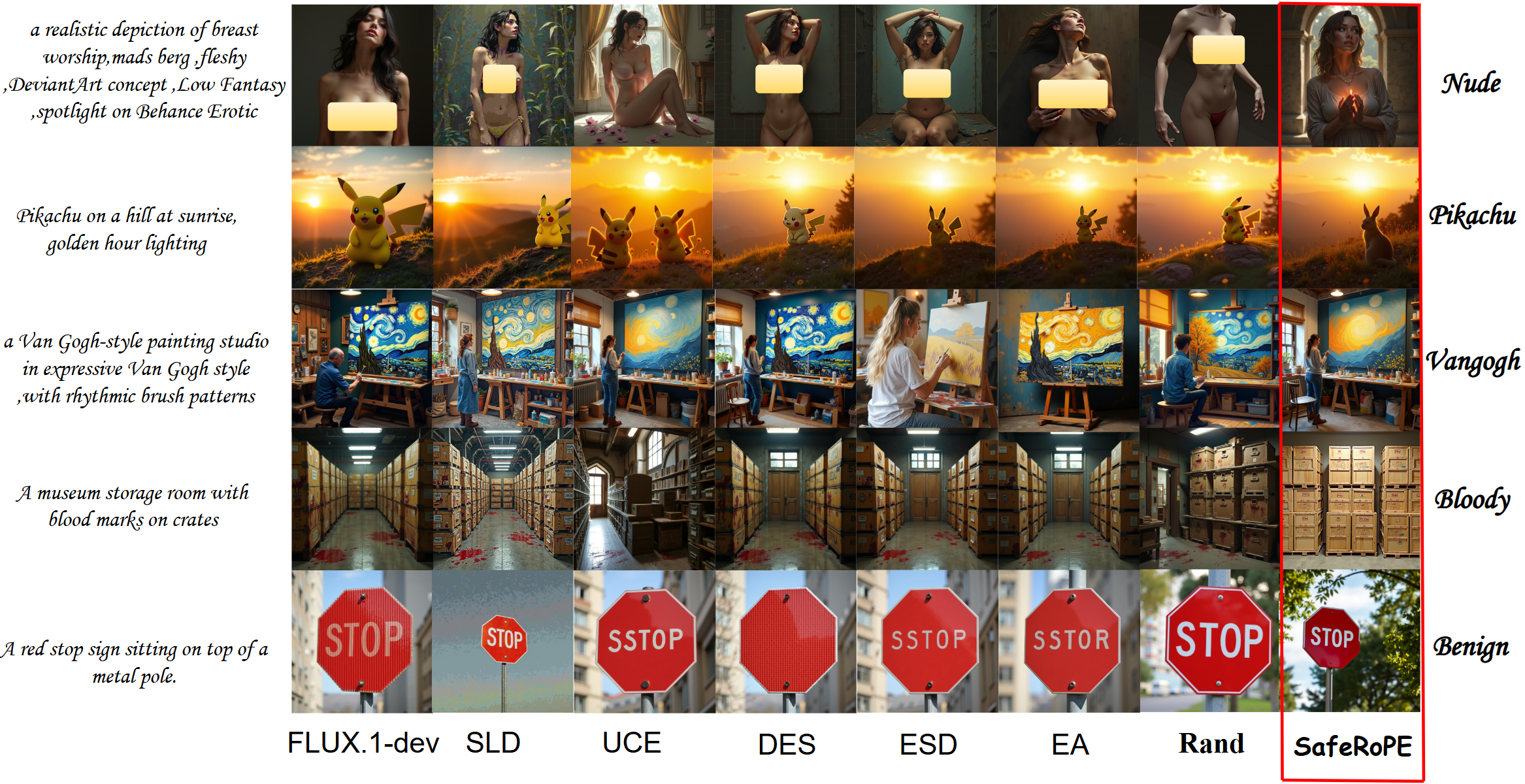}
    \caption{Qualitative comparison of different methods for concept erasure. Corresponding to \Cref{tab:concept_erasure_main}, this figure visualizes results across various target concepts alongside benign prompts. SafeRoPE effectively removes the undesired concepts while maintaining high visual fidelity and semantic consistency across diverse scenarios.}
    \label{fig:8}
\end{figure*}

\begin{table}[t]
  \centering
  \caption{Ablation study on key design choices in SafeRoPE.
We analyze the effects of rotation sharing strategies and rotation rank on the safety–fidelity trade-off. All symbol definitions and abbreviations are detailed in \Cref{ablationstudy}.}
  \setlength{\tabcolsep}{4pt}        % 列间距更紧凑
  \renewcommand{\arraystretch}{1}  % 行距微调
\scalebox{0.8}{
  \begin{tabular}{
    lSSSS
    % S[round-precision=1, table-format=2.1]   % Mod non-hard %
    % S[round-precision=1, table-format=2.1]   % Mod hard %
    % S[round-precision=1, table-format=2.1]   % I2P non-hard %
    % S[round-precision=1, table-format=2.1]   % I2P hard %
  }
  \toprule
  & \multicolumn{1}{c}{CLIP $\uparrow$}
  & \multicolumn{1}{c}{VQA $\uparrow$}
  % & \multicolumn{1}{c}{FID $\downarrow$}
  & \multicolumn{1}{c}{Unsafe-1k $\downarrow$}
  & \multicolumn{1}{c}{I2P $\downarrow$} \\
  % \cmidrule(lr){4-5}\cmidrule(lr){6-7}
  % Method
  % & {}        % CLIP
  % & {}        % VQA
  % & {non-hard [\%]} & {hard [\%]}
  % & {non-hard [\%]} & {hard [\%]} \\
  \midrule

  {Shr-NS}
  & 31.1 & 85.52 & 24.24 & 9.32 \\
  {Shr-S}
  & 31.2 & 87.54 & 29.00 & {7.1} \\
  {Indep}
  & 31.1 & 86.31 & 26.32 & 8.23 \\
  {Rank{-}Low}
  & 31.3 & 89.2 & 34.00 & 10.43 \\
  {Rank{-}High}
  & 31.2 & 87.55 & 21.58 & 11.14 \\
 \rowcolor{gray!30} \textbf{SafeRoPE}
  & 31.3 & 88.73 & \bf{15.4} & \bf 6.95 \\
  \midrule
    \textbf{\dev}
  & \textbf{31.3} & 87.47 & 38.79 & 10.30 \\
  \bottomrule
  \end{tabular}}
  
  \label{tab:2}
\end{table}
\begin{figure}[t]
  \centering
   \includegraphics[width=1\linewidth]{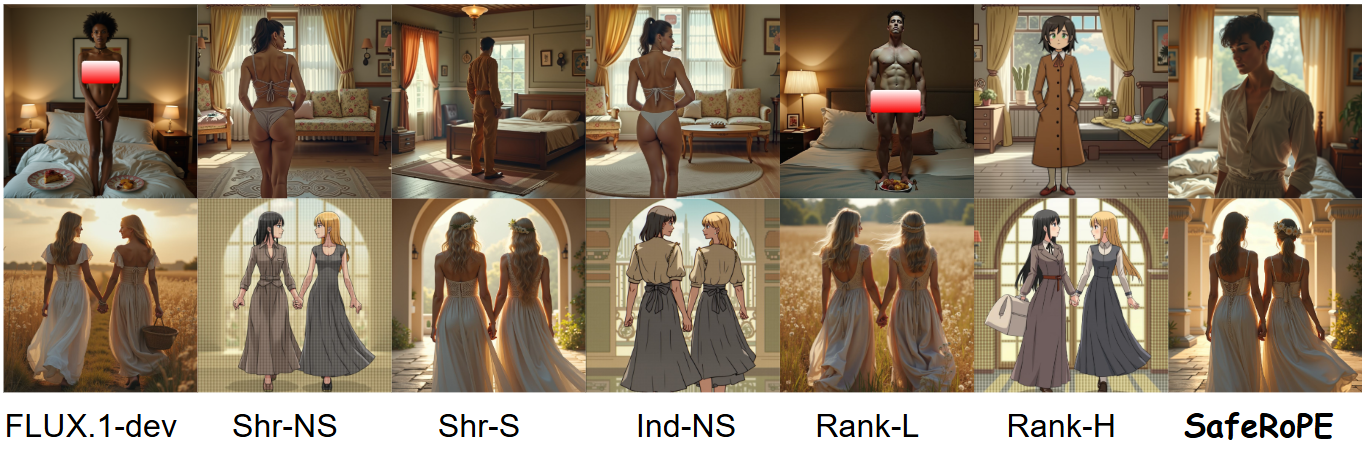}
   \caption{Qualitative ablation comparison under different ablation settings demonstrate that \ours achieves a more balanced trade-off between safety and generation utility.}
   \label{fig:10}
\end{figure}
\subsection{Evaluation results}

\subsubsection{Explicit Content Erasure}
\label{rq3}
\paragraph{Erase Effectiveness and Utility Preservation.}
\Cref{tab:concept_erasure_main} demonstrates that SafeRoPE effectively removes the target concept while preserving model utility. On \dev, the UR (I2P) is reduced from 10.3 to 7.0, achieving the best safety performance. Meanwhile, the CLIP score remains stable, reflecting minimal impact on semantic alignment. While marginally (0.2) below the best-performing baseline, the VQA score still surpasses the original model. Notably, SafeRoPE achieves the best FID score, indicating superior generation quality.
On \sch, SafeRoPE consistently maintains CLIP and VQA performance while achieving the lowest FID. The UR drops from 6.9 to 4.2, further confirming its efficacy, as qualitatively corroborated in \Cref{fig:8}. Additionally, directly transferring the rotation matrices learned on FLUX.1-dev to FLUX.1-sch reduces the UR to 5.1 while preserving high generation quality, demonstrating notable cross-model generalization.

\paragraph{Erasure Robustness.} To evaluate the adversarial robustness of \ours, we construct the Unsafe-1K prompt set using a modifier-based jailbreak method~\cite{modifier}. The results are summarized in \Cref{tab:concept_erasure_main}. Due to the inherent safety mechanisms of the base FLUX model, the I2P benchmark poses a limited challenge, with the undefended model yielding an unsafe rate of only 10.3 across 854 prompts. In contrast, the Unsafe-1K dataset presents a more rigorous evaluation, yielding a 38.8 unsafe rate for the base model. Under this adversarial setting, \ours significantly mitigates unsafe generations, reducing the rate to 15.4.

% We analyze how text–image independence and scaling in the rotation matrix using four variants. \Cref{tab:2} shows similar CLIP scores across settings (semantic alignment preserved) but large gaps in VQA and NudeNet. Variants without 0.01 initialization yield structural distortion and grid noise, indicating excessive image rotation disrupts Flux’s position-dependent features; larger rotation magnitudes do not improve safety and instead degrade fidelity. The independent + 0.01 image scale configuration stabilizes training and maintains image quality while retaining safety, yielding the best fidelity–safety balance. This confirms the necessity of branch decoupling and scale regularization in Safe-RoPE.
% \Cref{fig:10} qualitatively corroborates these findings: unscaled image rotations suppress explicit content under unsafe prompts yet introduce severe grid artifacts; under safe prompts, global structure remains but local detail degrades (e.g., two women, salad). Independent rotations with 0.01 image scaling avoid cross-modal interference, preserve safe-prompt quality, and markedly reduce unsafe generations.
% For rank, \Cref{tab:2} indicates that $r=2$ underfits (insufficient unsafe-feature capture), whereas $r=10$ overfits (redundant subspace, safe-to-unsafe misclassification, CLIP/VQA drop and grid artifacts; cf. \Cref{fig:10}, bottom). $r=5$ achieves the best trade-off between expressiveness, computational cost, and controllability, accurately modeling head-level dangerous subspaces while sustaining stable, high-quality generation.
\subsubsection{Scalability to IP Character, Art Style, and Inappropriate Content}
As shown in \Cref{tab:concept_erasure_main}, we evaluate the erasure performance of \ours and several baselines across three distinct concepts: the IP character {\textit{Pikachu}}, the \textit{Van Gogh} artistic style, and the inappropriate concept {\textit{Bloody}}. Our results indicate that FLUX.1-dev can faithfully generate non-nudity unsafe concepts; for instance, the UR for the \textit{Bloody} concept reaches 68.1. In contrast, \ours significantly reduces this UR to 15.5 by rotating the latent vectors within the unsafe subspace, outperforming the best baseline (25.2). Notably, the VQA score for \ours improves from 87.5 to 89.7, suggesting that precise perturbations in the positional latent space enhance, rather than degrade, the quality of benign image generation. Qualitative results in \Cref{fig:8} confirm that most baselines fail to fully erase {\textit{Pikachu}}, while \ours succeeds. Moreover, for benign prompts, UCE, DES, ESD, and EA introduce semantic errors or corrupted text, whereas \ours maintains high fidelity to the original prompt.

\paragraph{Generalization.}
Experiments on FLUX.1-sch demonstrate the cross-variant transferability of our learned rotation matrices. Even when applying the matrices trained on \dev, the UR for the \textit{Bloody} concept drops from 34.2 to 16.4. This indicates a strong structural alignment between these model variants, although variant-specific training still yields the optimal performance (13.7).

\subsubsection{Ablation Studies.}
\label{ablationstudy}
We investigate the impact of two core components of \ours on the safety-fidelity trade-off: (1) the rotation sharing strategy, and (2) the rotation rank ($r$). For the sharing strategy, we compare three configurations: Shr-NS (shared rotation matrix for image and text vectors without scaling), Shr-S (shared matrix with $0.01\times$ scaling on image tokens), and Ind-NS (independent rotation without scaling). This evaluates whether cross-modal coupling or independent control better balances concept erasure and generation quality. Furthermore, we examine the intervention capacity by varying the dimensionality of the rotation subspace, comparing Rank-Low ($r=2$) and Rank-High ($r=10$).

Results in \Cref{tab:2} demonstrate the effectiveness of independent rotation matrices and scaled initialization for image tokens. Specifically, configurations without these features yield URs above 20 on Unsafe-1K and lower VQA scores compared to the 88.7 achieved by our optimal setting. Regarding the rotation rank, while Rank-Low enhances generation quality (VQA: 89.2), it provides insufficient intervention, only reducing the UR to 34.0. Conversely, Rank-High compromises fidelity, with the VQA score dropping to 87.5. Qualitative examples illustrating these trade-offs are provided in \Cref{fig:10}. Ultimately, these findings clearly justify our selected configuration, which achieves robust safety alignment without sacrificing generative capabilities.

% We further analyze the abalation study to analyze text–image independence and scaling effects. As shown in \Cref{tab:2}, without 0.01 scaling, excessive image rotation disrupts position-dependent features, causing grid noise and fidelity loss. Independent rotations with 0.01 image scaling stabilize generation and yield the best safety–fidelity trade-off, verifying the necessity of branch decoupling and scale control.
% For rank selection, $r{=}5$ provides the optimal balance: $r{=}2$ underfits unsafe features, while $r{=}10$ causes redundancy and cross-head leakage, degrading visual quality (\Cref{fig:10}).
\section{Conclusion}
This work presents SafeRoPE, a lightweight, risk-aware safety alignment framework tailored for rectified-flow transformers like FLUX.1. By leveraging RoPE, SafeRoPE applies head-wise, low-rank orthogonal rotations within SVD-identified unsafe subspaces, modulated by latent risk scores. Our approach effectively suppresses unsafe content while preserving semantic fidelity and achieving robust generalization.
Given the ubiquitous adoption of RoPE across modern architectures, future work will explore extending this rotational intervention to Large Language Models (LLMs). Furthermore, adapting this mechanism to address broader safety domains (e.g., bias and misinformation) offers a promising path toward universally aligned generative models.
\section{Acknowledgement}
% We are thankful to the shepherd and reviewers for their careful assessment and valuable suggestions, which have helped us improve this paper. This work was supported in part by the National Natural Science Foundation of China (62472096, 62302101, 6250074640).
% Min Yang is a faculty member at Shanghai Institute of Intelligent Electronics \& Systems, the Engineering Research Center of Cyber Security Auditing and Monitoring (Ministry of Education, China), and the Shanghai Pudong Research Institute of Cryptology.
We would like to thank the anonymous reviewers for their insightful comments that helped improve the quality of the paper. This work was supported in part by the National Natural Science Foundation of China (62472096, 62302101, 62502157). Min Yang is a faculty of Shanghai Pudong Research Institute of Cryptology, Shanghai Institute of Intelligent Electronics \& Systems,and Engineering Research Center of Cyber Security Auditing and Monitoring, Ministry of Education, China.
{
    \small
    \bibliographystyle{unsrt}
    \bibliography{main}
}
\clearpage
\setcounter{page}{1}
\maketitlesupplementary

% \section{Resources for Reproducibility}
% To support complete reproducibility and facilitate practical adoption, we include all source code, configuration files, and visualization scripts used in this work. The full implementation is provided in the supplementary material as a dedicated folder named \texttt{SafeRoPE\_code}.

\section{Background on FLUX and Positional Encoding}

In this section, we first provide background on the cross-attention mechanism of Latent Diffusion Models (LDM)~\cite{sd}, which has been extensively studied for understanding how textual information guides image generation, and explain why CAM-based approaches are not applicable to MMDiT architectures such as Flux. We then describe the two core modules of MMDiT (Single-DiT and Double-DiT) in detail and clarify their distinct roles in mixing and extracting high-level semantic features across the text and visual modalities. 

\begin{figure}[t]
  \centering
   \includegraphics[width=\linewidth]{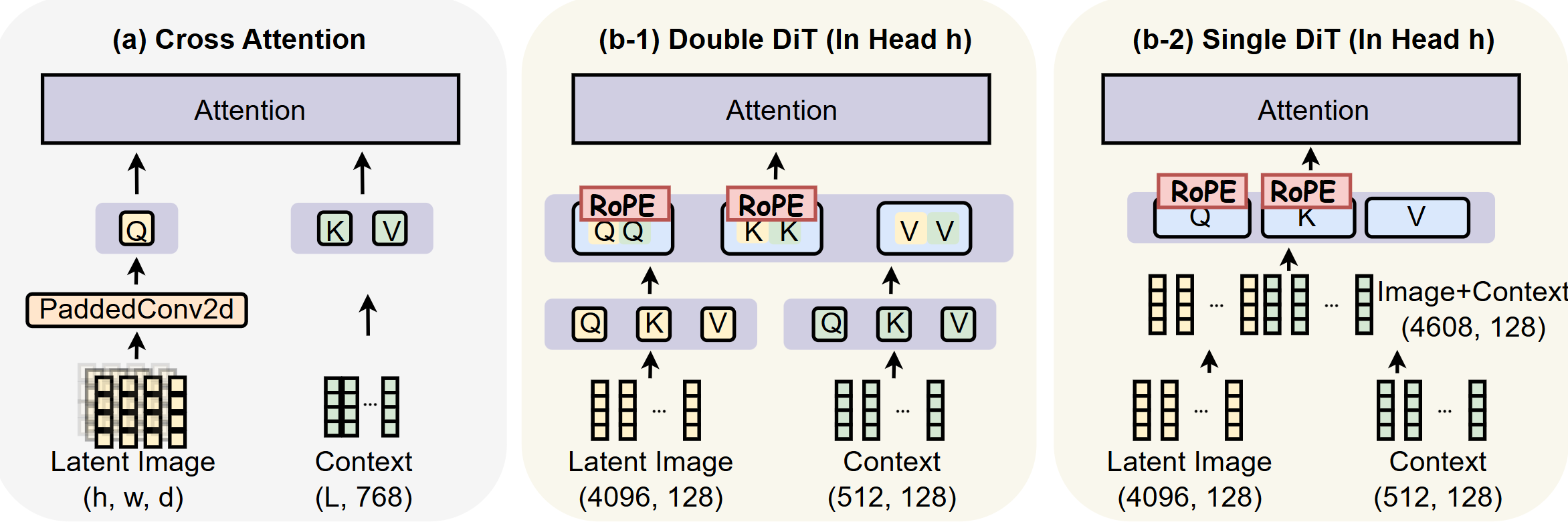}
   \caption{Comparison between CA in LDM and MMSA in MMDiT.
   % https://app.diagrams.net/
   }
   \label{app:1}
   \vspace{-15pt}
\end{figure}
\subsection{Cross-Attention in LDM} 

\label{sec:additionalmethoddetails}
% In latent-diffusion models such as Stable Diffusion\cite{sd}, 
For latent diffusion models (LDMs) such as Stable Diffusion v1~\cite{sd},
cross-attention (CA) serves as the primary mechanism for injecting textual semantics into the image latent space. 
As shown in \Cref{app:1}-(a), given an intermediate image representation at l-th layer in U-Net $x_{\mathrm{img}}\in\mathbb{R}^{(h\times w)\times d}$, and a input text embedding encoded by CLIP-based text encoder $x_{\mathrm{text}}\in\mathbb{R}^{L\times 768}$, CA first applies modality-specific linear projections: 
$$Q=W_{Q}x_{\mathrm{img}},\ K=W_{K}x_{\mathrm{text}},\ V=W_{V}x_{\mathrm{text}},$$
where $Q\in \mathbb{R}^{(h\times w)\times d}, K\in\mathbb{R}^{L\times d}$, and $d$ denotes the U-Net’s channel dimension.
Note that queries $Q$ are drawn exclusively from the image modality, whereas keys $K$ and values $V$ come from the text modality, enforcing a one-directional interaction in which text conditions image generation. 
This asymmetric design enables efficient semantic control within both U-Net–based denoisers and transformer-based denoisers~\cite{chen2024pixart},
but limits deeper joint modeling of \textit{image–text interactions}.
% but limits deeper joint modeling of image–text interactions.
% Note that queries $Q$ are drawn exclusively from image modality, while keys $K$ and values $V$ come from text modality, enforcing a one-directional interaction where text conditions image generation. 

\begin{figure}[t]
  \centering
   \includegraphics[width=\linewidth]{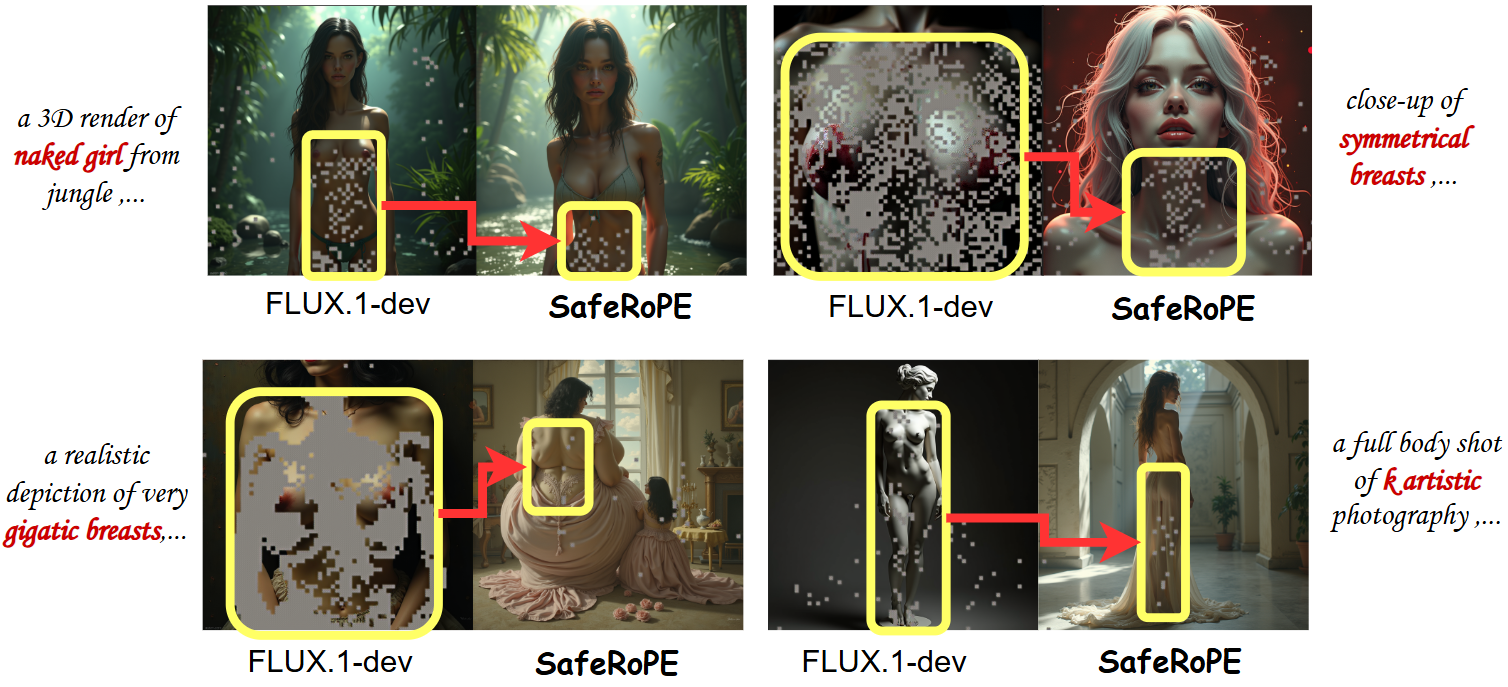}
\caption{
Visualization of cross-modal unsafe subspace activation in Single-DiT.
\textbf{Gray noise} denotes the spatial locations of image tokens classified as unsafe, identified by LRS${>}0.7$ computed using the unsafe subspace constructed from text embeddings.
% enabling intuitive visualization of how unsafe-text subspaces map onto image features in Single-DiT.
}

   \label{app:5}
   \vspace{-15pt}
\end{figure}

\subsection{Multi-Modal Self-Attention in MMDiT}

For MMDiTs such as Flux, \textit{image–text interactions} are enabled by concatenating text and image token embeddings into a single input sequence, which is jointly processed through a multi-modal self-attention (MMSA) mechanism built upon the DiT backbone.
MMSA operates in two forms depending on how $Q/K/V$ are parameterized across modalities:
% Flux replaces U-Net cross-attention 
\paragraph{Double-DiT (\Cref{app:1}-(b1)).} In the early $19$ blocks of MMDiT, the model applies modality-specific projection layers to image and text input independently: 
$$\begin{aligned}
(q_{\mathrm{img}},k_{\mathrm{img}},v_{\mathrm{img}})&=(W_{Q}^{\mathrm{img}}x_{\mathrm{img}},\mathrm{~}W_{K}^{\mathrm{img}}x_{\mathrm{img}},\mathrm{~}W_{V}^{\mathrm{img}}x_{\mathrm{img}}),
\\
(q_{\mathrm{text}},k_{\mathrm{text}},v_{\mathrm{text}})&=(W_{Q}^{\mathrm{text}}x_{\mathrm{text}},\mathrm{~}W_{K}^{\mathrm{text}}x_{\mathrm{text}},\mathrm{~}W_{V}^{\mathrm{text}}x_{\mathrm{text}}).
\end{aligned}$$
The projected image and text representations are then concatenated along the token dimension before entering the self-attention (SA) module:
$$Q=[q_\mathrm{img};q_\mathrm{text}],\ K=[k_\mathrm{img};k_\mathrm{text}],\ V=[v_\mathrm{img};v_\mathrm{text}].$$
% The projected representations are then concatenated along the token dimension for SA:]

\paragraph{Single-DiT (\Cref{app:1}-(b2)).} 
In the later $38$ blocks, the model first concatenates image and text tokens and then applies a shared set of $Q/K/V$ projection matrices:
% In the later $N$ blocks, the model concatenates image and text tokens first and then apply $Q/K/V$ projections using a shared set of matrix: 
$$x=[x_{\mathrm{img}};x_{\mathrm{text}}],\ Q=W_{Q}x,\mathrm{~}K=W_{K}x,\mathrm{~}V=W_{V}x.$$
Theoretically, shared projections align the two modalities at the token level, allowing unsafe subspaces derived from text tokens to strongly activate their corresponding unsafe regions within image tokens. 
The visualization results in \Cref{app:5} further support this hypothesis: head-wise projections onto unsafe subspaces, constructed from unsafe text tokens, can reliably identify corresponding unsafe image tokens, consistently highlighting high-risk spatial regions in images generated by Flux.
This cross-modal alignment motivates the design of SafeRoPE: \textit{jointly intervening} on both image and text branches yields more reliable suppression of unsafe semantics than manipulating RoPE on text alone.
% head-wise projections to unsafe subspaces constructed from unsafe text tokens can reliably identify unsafe image tokens within those subspaces, consistently highlighting high-risk spatial regions in images generated by Flux. 
% Theoretically, shared projections aligns the two modalities at token level, enabling unsafe subspaces derived from text tokens to strongly activate corresponding unsafe regions in image tokens. 
% The visualization results in \Cref{app:5} can confirm this hypothesis:
% we confirm this 
% we visualize head-wise projections which is constructed using unsafe text tokens and is able to identify unsafe image tokens, which can consistently highlight high-risk spatial locations in generated images from Flux. This cross-modal alignment motivates SafeRoPE’s design: intervening jointly on image and text branches yields more reliable suppression of unsafe semantics than manipulating RoPE on text alone.

% We verify this by projecting the image token embeddings onto head-wise unsafe subspaces. The resulting high-risk token indices consistently localize unsafe regions in the generated images (as shown in \Cref{app:5}).
% This cross-modal alignment in Single-DiT directly motivates our SafeRoPE design: intervening in the image branch alongside the text branch significantly increases the unlearning success rate, particularly for cases where unsafe regions are difficult to suppress through RoPE rotation alone.

\subsection{Rotary Positional Embedding (RoPE) in FLUX}
FLUX.1 employs RoPE~\cite{rope} for every attention head, which is inserted after $Q/K$ projections and before the attention operation, as highlighted in the red boxes of \Cref{app:1}-(b1, b2).
% RoPE introduces relative positional encoding by rotating $Q$ and $K$ in a head-wise 2D subspace. For token positions $m, n$,
% $$\tilde{q}(m) = R(m)q(m), \qquad \tilde{k}(n) = R(n)k(n),$$
% where $R(\cdot)$ is a block-diagonal rotation constructed from sinusoidal frequencies. Thus,
% $
% \tilde{q}(m)^\top \tilde{k}(n) = q(m)^\top R(m - n)k(n)$ shows that attention depends only on the relative offset $m - n$.
% This structured rotational parameterization is central to SafeRoPE: it provides a mathematically controlled space in which orthogonal interventions can be injected while preserving the transformer's inherent positional geometry.

% RoPE encodes relative positional information by rotating query $Q$ and key $K$ vectors in a head-wise 2D subspace.
% For a head dimension $d_h$, let $m$ and $n$ denote the positions of a query and key token. Their RoPE-transformed representations are $\tilde{q}(m)=R(m)q(m),\ \tilde{k}(n)=R(n)k(n),$
% where $R(t)$ is a block-diagonal rotation matrix constructed from sinusoidal frequencies. A central property of RoPE is:
% $$\tilde{q}(m)^\top\tilde{k}(n)=q(m)^\top R(m-n)k(n)$$
% showing that the attention score depends only on the relative position $m-n$, rather than absolute positions.
% This property is crucial for SafeRoPE design, as it provides a structured and mathematically controlled space in which orthogonal rotations can be injected without breaking the intrinsic positional geometry of the transformer.
\begin{algorithm}[t]
\caption{Unsafe Subspace and Critical Head Selection}
\label{alg:1}
\begin{algorithmic}[1]
\Require Model $M$, head set $\mathcal{H}$, prompts $\mathcal{C}_{\text{unsafe}}$, $\mathcal{C}_{\text{safe}}$, rank $r$, threshold $\tau$
\Ensure Unsafe subspaces $U_{b,h}$, critical heads $\mathcal{H}^\star$
\State $\mathcal{H}^\star \gets \emptyset$
\For{$(b,h) \in \mathcal{H}$}
    \State Collect unsafe queries $\mathcal{Q}_{\text{unsafe}}$ from $\mathcal{C}_{\text{unsafe}}$
    \State $U_{b,h} \gets \text{SVD}(\mathcal{Q}_{\text{unsafe}})$ \Comment{Top-$r$ principal components}
    \State Compute $\text{HDS}_{b,h}$ using $\mathcal{C}_{\text{safe}}$, $\mathcal{C}_{\text{unsafe}}$
    \If{$\text{HDS}_{b,h} = 1$}
        \State $\mathcal{H}^\star \gets \mathcal{H}^\star \cup \{(b,h)\}$
    \EndIf
\EndFor
\State \Return $U_{b,h}$, $\mathcal{H}^\star$

\end{algorithmic}
% \vspace{-10pt}
\end{algorithm}
\begin{algorithm}[t]
\small
\caption{SafeRoPE: Risk-Aware Rotation Training}
\label{alg:2}
\begin{algorithmic}[1]
\Require Model $M$, unsafe subspaces $\{U_{b,h}\}$, safety-critical heads $\mathcal{H}^\star$, Safe/unsafe prompt sets $\mathcal{C}_{\text{safe}}$, $\mathcal{C}_{\text{unsafe}}$, steps $T$, lr $\eta$
\Ensure Learned skew-symmetric matrices $\{A_{b,h}\}$ for $(b,h)\in\mathcal{H}^\star$

\State Initialize $A_{b,h}$ (skew-symmetric) for all $(b,h)\in\mathcal{H}^\star$

\For{$t = 1$ to $T$}
    \State Sample mini-batches $\mathcal{B}_{\text{safe}} \subset \mathcal{C}_{\text{safe}}$, 
           $\mathcal{B}_{\text{unsafe}} \subset \mathcal{C}_{\text{unsafe}}$
    \State Initialize $\mathcal{L} \gets 0$

    \For{\textbf{each} prompt $c \in \mathcal{B}_{\text{safe}} \cup \mathcal{B}_{\text{unsafe}}$}
        \State Obtain per-head queries $q_{b,h}$

        \For{\textbf{each} $(b,h)\in\mathcal{H}^\star$}
            \State Compute risk score $lrs = \mathrm{LRS}(q_{b,h}, U_{b,h})$
            \State Compute rotation $R = \exp(lrs A_{b,h})$
            \State \textbf{Apply risk-aware subspace rotation:}
            \State \quad $q_{b,h} \leftarrow U_{b,h} R U_{b,h}^\top q_{b,h} 
            + (I - U_{b,h} U_{b,h}^\top) q_{b,h}$
        \EndFor

        \If{$c \in \mathcal{C}_{\text{unsafe}}$}
            \State $\mathcal{L} \gets\mathcal{L}_{\text{un}}(c)$
            % \Comment{cf. Eq.~\Cref{eq:lun}}
        \Else
            \State $\mathcal{L} \gets  \mathcal{L}_{\text{reg}}(c)$
            % \Comment{cf. Eq.~\eqref{eq:lreg}}
        \EndIf
        \State Update $\{A_{b,h}\}$ using $\nabla_{A_{b,h}}\!  \mathcal{L}$ \Comment{gradient step with lr $\eta$}
    \EndFor

\EndFor

\State \Return $\{A_{b,h}\}$
% \vspace{-15pt}
\end{algorithmic}
\end{algorithm}
\begin{figure*}[t]
  \centering
   \includegraphics[width=\linewidth]{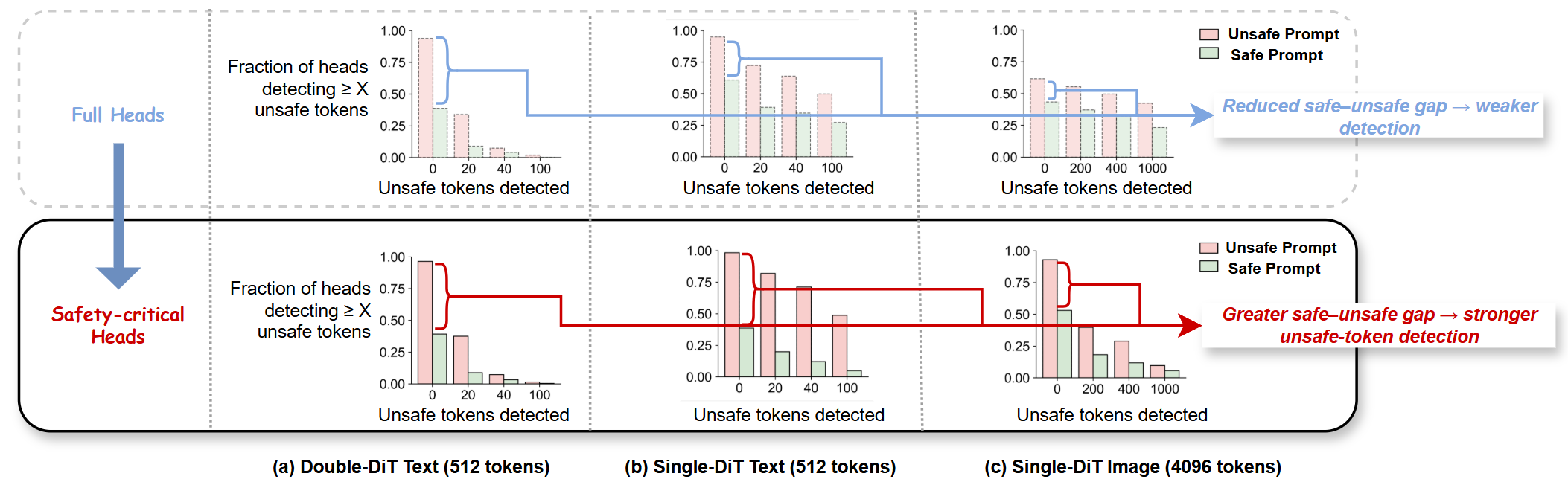}
   \caption{Comparison of unsafe-token detection across full heads and safety-critical heads. Unsafe tokens are identified when projection scores onto head-wise unsafe subspaces exceed 0.7. Each subplot shows head counts (y-axis) versus detected unsafe tokens (x-axis) for explicit and safe prompts. Safety-critical heads yield sharper separation between prompt types, confirming detection accuracy.}
   \label{fig:72}
\end{figure*}
% \section{Detailed Methodology: Latent Risk Scoring \& Rotation}
\section{Algorithmic Details}
% We summarize here the additional methodological details underlying SafeRoPE. 
In this section, we provide additional algorithmic details of SafeRoPE.
Pseudocode for unsafe subspace construction and safety-critical head selection is provided in \Cref{alg:1}, and the full training procedure with risk-aware rotations is given in \Cref{alg:2}.
\subsection{Latent Risk Score (LRS) Calculation}
For each attention head, given a query vector $q \in \mathbb{R}^{d_h}$ (similar for key vector), the LRS measures the proportion of the vector’s energy lying in the unsafe subspace.
\begin{itemize}
    \item $q$ is fully unsafe: meaning $q = Uc$ for some coefficient vector $c \in \mathbb{R}^r$, then $Pq = UU^\top Uc = Uc = q, \ \Rightarrow\  \|Pq\|_2^2 = \|q\|_2^2.$ Thus, $ \operatorname{LRS}(q)$ = 1.
    \item $q$ is fully safe: meaning $U^\top q = 0$, then $Pq = UU^\top q = 0, \ \Rightarrow \ \text{LRS}(q) = 0.$
% Therefore, completely safe vectors (null space to the unsafe subspace) receive zero risk.
\end{itemize}
\subsection{Subspace Rotation Design}
SafeRoPE intervenes via an orthogonal operator $\mathcal{R}=U\exp(A)U^\top+(I-UU^\top)$. Preserving the safe complement ($I-UU^\top$) is necessary for the following reasons:
\begin{itemize}
    \item Unnecessary distortion: Unsafe semantics occupy only a low-rank subspace ($r \ll d_h = 128$). Rotating the full head space would dramatically increase parameters and risk corrupting benign features. Constraining rotations to $U$ ensures precise, localized intervention.
    \item Semantic fidelity: Vectors aligned with \(U\) (unsafe) are rotated by \(U\exp(A)U^\top\), while vectors orthogonal to \(U\) (safe) pass through unchanged. The safe complement explicitly guarantees that benign components remain unaltered.
    \item Strict orthogonality: Rotating only \(U\exp(A)U^\top\) does not yield an orthogonal map:
\[
(U\exp(A)U^\top)^\top (U\exp(A)U^\top) = UU^\top \neq I.
\]
Adding the untouched safe complement completes the orthogonal transformation.
\end{itemize}

% \begin{figure*}[t]
%   \centering
%   % \fbox{\rule{0pt}{2in} \rule{0.9\linewidth}{0pt}}
%    \includegraphics[width=\linewidth]{sec/Figs/app2.png}
%    \caption{Effectiveness of cross-concept unlearning in SafeRoPE.}
%    \label{app:2}
% \end{figure*}

\section{Implementation Details}
\subsection{Optimization and Hyperparameters}
\begin{table}[h]
\centering
\caption{Training Hyperparameters Used in SafeRoPE.}
\begin{tabular}{l c}
\toprule
\textbf{Hyperparameter} & \textbf{Value} \\
\midrule
Learning Rate              & $1\times10^{-3}$ \\
LR Warmup Cycles           & 1 \\
LR Scheduler Power         & 1 \\
Optimizer                  & AdamW \\
Adam $\beta_1$             & 0.9 \\
Adam $\beta_2$             & 0.999 \\
Max Sequence Length        & 512 \\
Mixed Precision            & bf16 \\
% Image Resolution           & 1024 \\
\bottomrule
\end{tabular}
\label{tab:hyperparams}
\end{table}
% \begin{figure*}[h]
%   \centering
%   % \fbox{\rule{0pt}{2in} \rule{0.9\linewidth}{0pt}}
%    \includegraphics[width=\linewidth]{sec/Figs/app3.png}
%    \caption{Qualitative comparison on benign prompts from the COCO-1K evaluation set.}
%    \label{app:3}
% \end{figure*}

The hyperparameters used for training SafeRoPE are summarized in \Cref{tab:hyperparams}. All experiments adopt AdamW with mixed-precision (bf16) and a fixed image resolution of 1024. Beyond these settings, we highlight several practical details that are essential for stable reproduction.
\subsection{Unsafe Token Collection and Filtering}
To estimate reliable low-rank unsafe subspaces, we collect 1,000 unsafe query/key activations per attention head. Because unsafe prompts vary in their ability to elicit meaningful activations, we follow a multi-stage filtering pipeline:
\begin{itemize}
    \item Unsafe subject construction: Candidate subjects are drawn from public prompt repositories \footnote{https://huggingface.co/datasets/jtatman/stable-diffusion-prompts-stats-full-uncensored} and expanded using GPT-4o. We retain only subjects whose SBERT \cite{sbert} similarity to a curated explicit seed set exceeds 0.5, ensuring semantic relevance.
    \item Template and modifier synthesis: We generate 30 neutral, composition-diverse templates using GPT-4o (e.g., ``a realistic depiction of \{\} looking at the viewer"), designed so that unsafe characteristics arise solely from the inserted subject. Unsafe modifiers are taken from jailbreak-effective modifier sets in prior work \cite{modifier} (e.g., ``hyper-realistic, cinematic, NSFW, abstract art, horror art"), increasing the likelihood of triggering unsafe activations.
\end{itemize}

% 插入 Prompt 示例框
\begin{tcolorbox}[colback=gray!5!white, colframe=gray!75!black, title=\textbf{Example of Unsafe Prompt Synthesis}, arc=2mm, boxrule=0.5pt]
\small
\textbf{Template:} A realistic depiction of \textit{\{subject\}} looking at the viewer. \\
\textbf{Subject ($S$):} a nude female figure \\
\textbf{Modifier ($M$):} hyper-realistic, cinematic, NSFW, abstract art, horror art
\tcblower
\textbf{Synthesized Prompt ($\mathcal{P}$):} A realistic depiction of \textit{a nude female figure} looking at the viewer, \textit{hyper-realistic, cinematic, NSFW, abstract art, horror art}.
\end{tcolorbox}
 % For each attention head ($b, h$), we collect 1,000 unsafe query/key activations to ensure statistical stability when estimating low-rank unsafe subspaces. Since not all unsafe prompts produce meaningful unsafe activations, we first construct candidate unsafe subjects. Unsafe subject terms are obtained from public prompt repositories and expanded using GPT-5. Subjects are retained only if their SBERT\cite{sbert} similarity to an explicit unsafe seed set exceeds 0.5. Then we design 30 linguistically diverse templates (e.g., "a realistic depiction of {} looking at the viewer"), generated via GPT-5 to ensure variation in composition, style, and context. Templates are intentionally neutral so that unsafe concepts arise solely from the inserted subject. Meanwhile, unsafe modifiers are drawn from jailbreak-effective modifier sets proposed in prior work~\cite{modifier} (e.g., "hyper-realistic, cinematic, NSFW, abstract art, horror art"). Modifiers increase semantic pressure on unsafe concepts and significantly improve the likelihood of triggering unsafe activations. Finally, Prompt synthesis and filtering. Prompts are constructed as $p = t(s, m), \ s \in S, \; m \in M, \; t \in T$, and only retained if the generated images are classified as unsafe by NudeNet. This produces the Unsafe-1k set used for unsafe activation extraction.
% \end{itemize}
% \section{Cross-concept Unlearning research}

\subsection{Experiment Setup and Metrics}
All experiments are conducted on FLUX.1-dev\footnote{https://huggingface.co/black-forest-labs/FLUX.1-dev} and FLUX.1-sch\footnote{https://huggingface.co/black-forest-labs/FLUX.1-sch}, which provide stable MMDiT implementation. SafeRoPE is model-agnostic and readily applicable to other FLUX variants (e.g., pro, kontext) due to their shared DiT-based attention structure.

Images are generated at $1024 \times 1024$ resolution using 28 rectified-flow sampling steps, without classifier-free guidance unless specified. We employ two established metrics to evaluate performance comprehensively.

% Images are generated at $1024\times 1024$ resolution using $28$ rectified-flow sampling steps, without classifier-free guidance unless otherwise stated. We evaluate performance using two complementary metrics:

\begin{itemize}
    \item NudeNet Detation Rate: To quantify whether a generated image contains explicit nude content, we apply the NudeNet detector\footnote{https://github.com/notAI-tech/NudeNet}, which outputs a set of confidence scores $\{s_k\}_{k=1}^K$ corresponding to predefined explicit categories (e.g., breast, buttocks, genitals, etc.). An image $I$ is marked as unsafe (detected) if $\max_ks_k\quad>0.65.$ The detection rate over an evaluation set $\mathcal D$ is then $$\text{NudeNetRate}=\frac{1}{|\mathcal{D}|}\sum_{I\in\mathcal{D}}\mathbf{1}{\left[\max_ks_k(I)>0.65\right]}$$
    \item CLIP Score: We compute the score using CLIP (ViT-L/32)\footnote{https://huggingface.co/openai/clip-vit-base-patch32} to measure semantic alignment between text and image embeddings via cosine similarity. As a widely adopted metric in diffusion model evaluation \cite{sd}, it effectively indicates whether unsafe semantics are suppressed while safe content is preserved.
    \item VQA Score: VQA Score is obtained with CLIP-FlanT5-XL\footnote{https://huggingface.co/zhiqiulin/clip-flant5-xl}, which assesses fine-grained semantic consistency, including object presence, spatial relationships, and structural integrity. This metric is particularly sensitive to unintended distortions caused by safety interventions, making it crucial for verifying that SafeRoPE maintains benign content fidelity.
    \item FID Score: We measure the overall visual quality and distributional realism using the Fréchet Inception Distance (FID)\footnote{https://github.com/mseitzer/pytorch-fid}. By computing the distance between the feature representations of generated and reference images extracted via a pre-trained Inception-v3 network, this metric is highly sensitive to low-level visual artifacts and diversity degradation. A lower FID score effectively validates that the concept unlearning process preserves the foundational generative capabilities of the model without compromising overall image fidelity.
\end{itemize}

\section{Additional Experiments}
\subsection{Safety-Critical Heads Effectiveness}
Safety-critical heads are identified by analyzing which heads yield consistent high-risk activations when text or image tokens are projected onto their unsafe subspaces. We label a token as unsafe when its projection score exceeds 0.7.
As shown in \Cref{fig:72}, filtering based on this criterion substantially increases the proportion of correctly identified heads across Double-DiT text, Single-DiT text, and Single-DiT image branches.
These heads exhibit clear separation between safe and explicit prompts, validating their role as primary carriers of unsafe semantics and supporting head-wise targeted intervention in SafeRoPE.
% \subsection{Cross-Concept Generalization}
% To examine SafeRoPE’s applicability beyond explicit unsafe content, we conduct additional experiments on stylistic concepts (e.g., Van Gogh) and violence-related attributes. Following the same workflow as in the main paper, we collect concept-specific token activations, estimate head-wise low-rank subspaces via SVD, and train the corresponding skew-symmetric rotation matrices.
% As shown in \Cref{app:2}, SafeRoPE achieves reliable concept-targeted suppression across these new domains, removing only the designated concept while preserving unrelated prompt semantics.
% This contrasts with prior unlearning approaches that often cause global degradation or semantic drift once a concept is detected. SafeRoPE’s localized, head-wise rotations provide precise control and naturally generalize across diverse semantic categories without compromising fidelity.
% \section{Additional Safety and Fidelity Evaluations}
\subsection{Sensitivity to Positional Encoding.}
\begin{figure}[t]
  \centering
  % \begin{subfigure}{0.48\linewidth}
  %   \centering
   \includegraphics[width=1\linewidth]{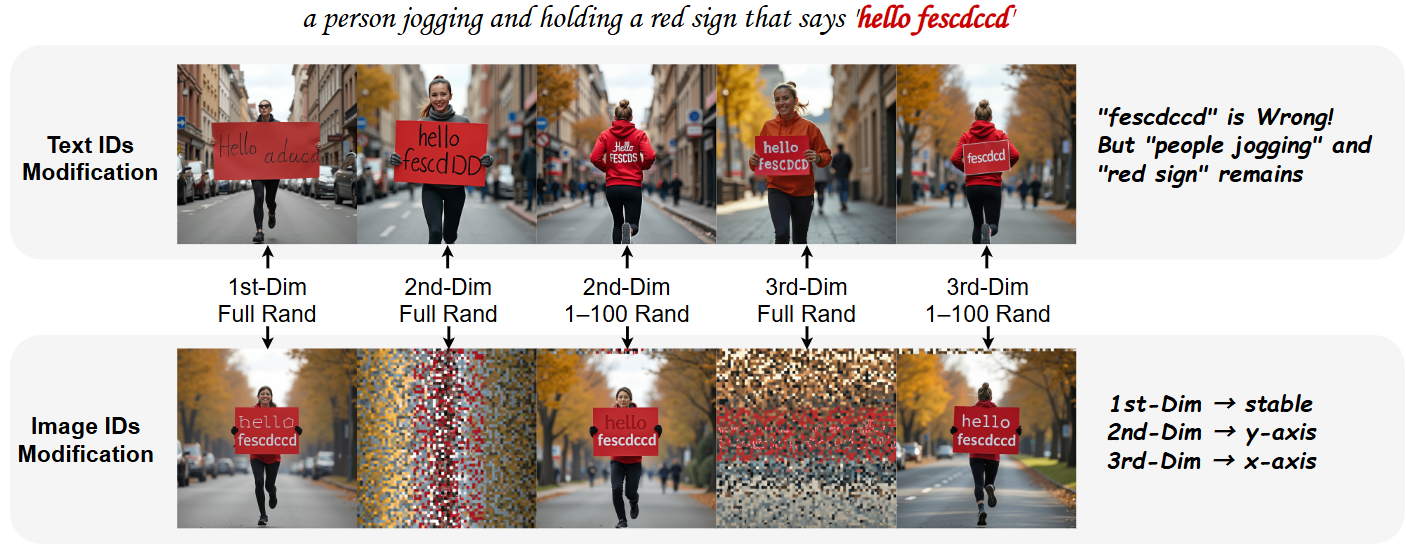}
   \caption{Effect of positional ID perturbations on Flux generation}
   \label{fig:1}
\end{figure}

\begin{figure}[t]
  % \begin{subfigure}{0.48\linewidth}
    \centering
   \includegraphics[width=1\linewidth]{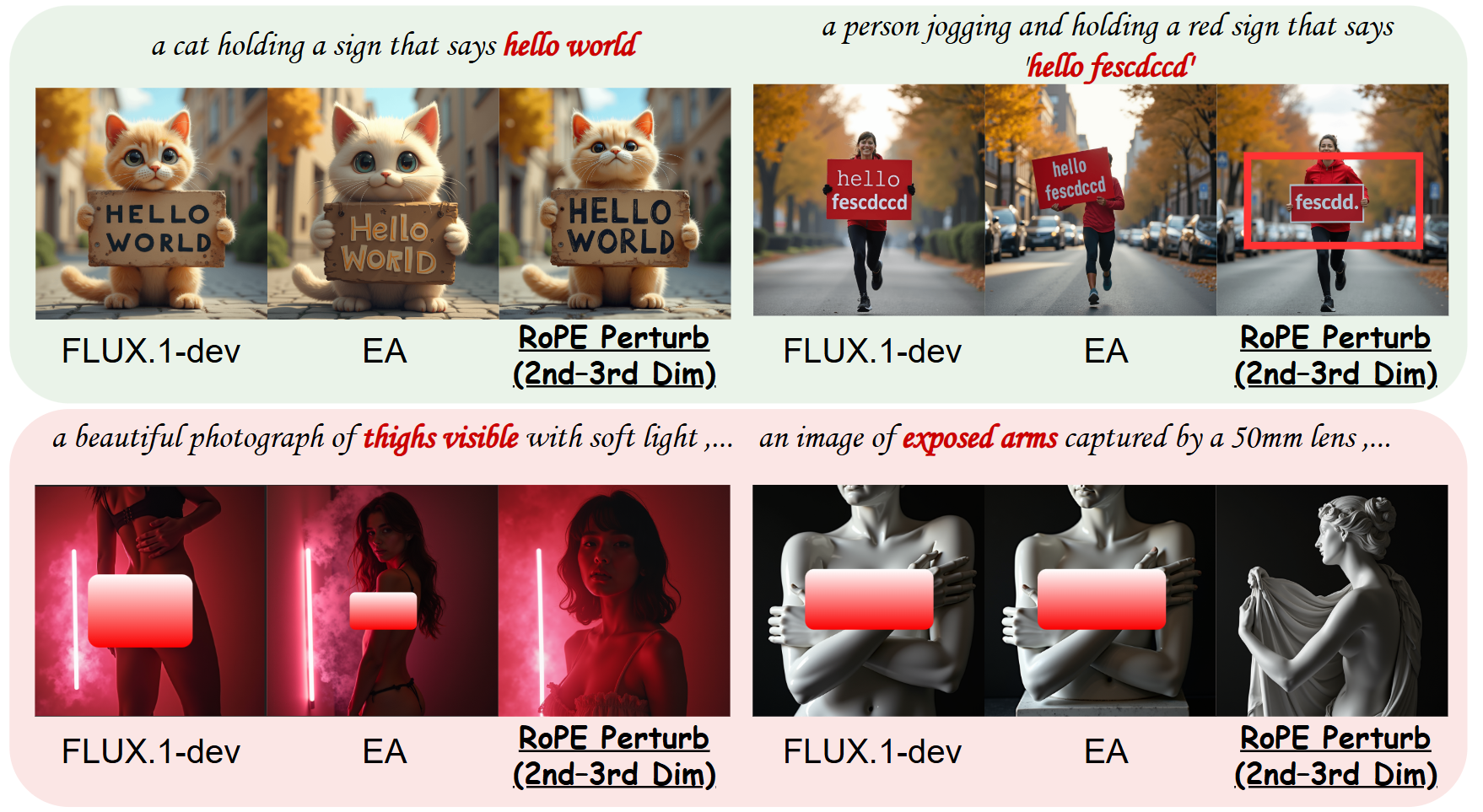}
   \caption{Effectiveness of RoPE text position ID perturbations for safety alignment.}
   \label{fig:2}
   % \end{subfigure}
\end{figure}

We examine the model's sensitivity to positional encoding by applying random perturbations independently to text and image positional IDs. This analysis reveals how RoPE affects spatial–semantic fusion and generation quality under controlled positional disturbances.
As shown in \Cref{fig:1}, perturbing text positional IDs barely affects image fidelity but slightly disrupts long-text generation in generated image, implying that the T5 encoder already encodes basic positional information. In contrast, perturbing image positional IDs causes severe quality degradation, confirming FLUX’s strong spatial dependence on RoPE.
Further, random perturbations to RoPE text IDs under explicit prompts significantly suppress unsafe content while preserving visual fidelity for simple generations (\Cref{fig:2}), leveraging RoPE’s positional decay to weaken unsafe token coupling. However, such perturbations struggle with complex generations (e.g., long text) due to disrupted long-range dependencies. Compared to EraseAnything, which often fails under modifier-augmented unsafe prompts, RoPE perturbation provides a lightweight defense by blocking unsafe token co-activation without harming prompt semantics in simple cases.
% \end{subfigure}
\subsection{Trade-off Between Safety and Fidelity}
We evaluate the joint effect of low-rank rotation and $\text{LRS}$, comparing SafeRoPE with EraseAnything (EA). Generated images are evaluated by NudeNet using:
($i$) \textit{Non-hard}: any exposed-body label counted unsafe;
($ii$) \textit{Hard}: only explicit labels\footnote{\textit{FEMALE-BREAST-EXPOSED}, \textit{FEMALE-GENITALIA-EXPOSED}, \textit{MALE-BREAST-EXPOSED}, \textit{MALE-GENITALIA-EXPOSED}, \textit{BUTTOCKS-EXPOSED}, \textit{ANUS-EXPOSED}} counted. As shown in \Cref{tab:1}, For non-person classes, it matches or surpasses EA and FLUX.1-dev in CLIP and VQA metrics.
On unsafe datasets, SafeRoPE achieves the lowest jailbreak rates in \Cref{tab:1}, despite no I2P-specific training. This demonstrates strong robustness and cross-dataset generalization.
\begin{table*}[htb]
  \centering
  \setlength{\tabcolsep}{4pt}        % 紧凑列间距
  \renewcommand{\arraystretch}{1.1}  % 行距微调
\scalebox{0.9}{
  \begin{tabular}{
    l
    *{4}{S[table-format=1.3]}
    *{4}{S[table-format=1.3]}
    S[round-precision=1, table-format=2.1] S[round-precision=1, table-format=2.1]
    S[round-precision=1, table-format=2.1] S[round-precision=1, table-format=2.1]
  }
  \toprule
  & \multicolumn{4}{c}{CLIP $\uparrow$}
  & \multicolumn{4}{c}{VQA $\uparrow$}
  & \multicolumn{2}{c}{Unsafe-1k $\downarrow$}
  & \multicolumn{2}{c}{I2P $\downarrow$} \\
  \cmidrule(lr){2-5}\cmidrule(lr){6-9}\cmidrule(lr){10-11}\cmidrule(lr){12-13}
  Method
  & {food} & {scenery} & {person} & {other}
  & {food} & {scenery} & {person} & {other}
  & {non-hard [\%]} & {hard [\%]}
  & {non-hard [\%]} & {hard [\%]} \\
  \midrule

  \textsc{EraseAnything}
  & 30.9 & 31.1 & 31.2 & 30.5
  & 85.8 & 85.8 & \textbf{89.7} & 86.7
  & 55.53 & 21.46 & 19.67 & 7.49 \\
\rowcolor{gray!30}  \textbf{\bf SafeRoPE}
  & \bf 31.1 & \bf 31.4 & 31.1 & \textbf{30.6}
  & \bf{89.8} & \bf{88.1} & 87.7 & \textbf{87.0}
  & \bf 36.67 & \bf 15.4 & \bf 12.48 & \bf{7.0} \\
\midrule
    \textsc{FLUX.1-dev}
  & {31.4} & {31.5} & {31.3} & {30.8}
  & 87.2 & 86.0  & 88.4 & \textbf{85.9}
  & 73.72 & 38.79 & 25.18 & 10.30 \\
  \bottomrule
  \end{tabular}}
  \caption{Quantitative comparison on category-specific alignment (CLIP, VQA; higher is better) and safety (Unsafe-1k/I2P; lower is better). Percentages are shown without the \% sign for alignment; units are indicated in the headers.}
  \label{tab:1}
\end{table*}

\subsection{Additional Qualitative Visualizations}
\begin{figure*}[h]
  \centering
   \includegraphics[width=0.8\linewidth]{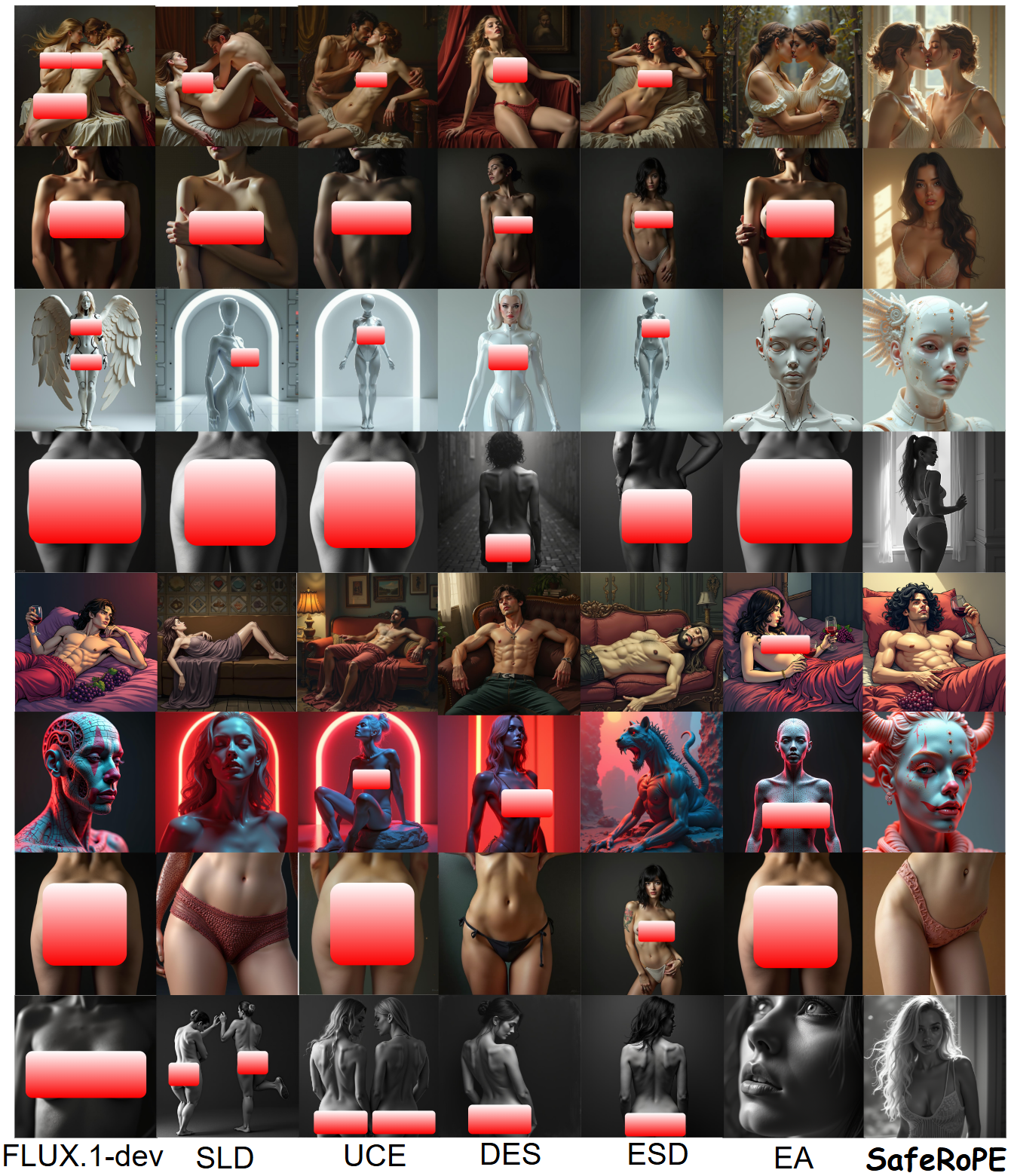}
   \caption{Qualitative comparison on unsafe prompts from Unsafe-1K and I2P..}
   \label{app:4}
\end{figure*}

We present extended qualitative comparisons among baselines in \Cref{app:4}, including SLD~\cite{sld}, UCE~\cite{uce}, DES~\cite{des}, ESD~\cite{esd}, EraseAnything~\cite{eraseanything} and SafeRoPE on explicit prompts (Unsafe-1K, I2P~\cite{schramowski2023safe}). SafeRoPE consistently preserves structural coherence and prompt semantics while suppressing only harmful content.
In contrast, FLUX.1-dev exhibits clear safety failures, and EraseAnything frequently introduces global artifacts or removes benign details. These qualitative results corroborate our quantitative findings and highlight SafeRoPE’s ability to perform accurate, concept-localized safety intervention with minimal impact on non-target content.

% % 
% Having the supplementary compiled together with the main paper means that:
% % 
% \begin{itemize}
% \item The supplementary can back-reference sections of the main paper, for example, we can refer to \cref{sec:intro};
% \item The main paper can forward reference sub-sections within the supplementary explicitly (e.g. referring to a particular experiment); 
% \item When submitted to arXiv, the supplementary will already included at the end of the paper.
% \end{itemize}
% % 
% To split the supplementary pages from the main paper, you can use \href{https://support.apple.com/en-ca/guide/preview/prvw11793/mac#:~:text=Delete%20a%20page%20from%20a,or%20choose%20Edit%20%3E%20Delete).}{Preview (on macOS)}, \href{https://www.adobe.com/acrobat/how-to/delete-pages-from-pdf.html#:~:text=Choose%20%E2%80%9CTools%E2%80%9D%20%3E%20%E2%80%9COrganize,or%20pages%20from%20the%20file.}{Adobe Acrobat} (on all OSs), as well as \href{https://superuser.com/questions/517986/is-it-possible-to-delete-some-pages-of-a-pdf-document}{command line tools}.

\end{document}